%% file: main.tex
\documentclass[CSUR,acmlarge]{acmart}

\AtBeginDocument{%
  \providecommand\BibTeX{{%
    \normalfont B\kern-0.5em{\scshape i\kern-0.25em b}\kern-0.8em\TeX}}}

\setcopyright{acmcopyright}
\acmJournal{CSUR}
\copyrightyear{2023}
\acmYear{2018}
\acmDOI{XXXXXXX.XXXXXXX}

\acmJournal{CSUR}
\acmVolume{1}
\acmNumber{1}
\acmArticle{1}
\acmMonth{1}

\acmPrice{15.00}
\acmISBN{978-1-4503-XXXX-X/18/06}

\usepackage{amsmath}
\usepackage{balance}       
\usepackage{graphics}      
\usepackage{graphicx}
\usepackage{adjustbox}
\usepackage[T1]{fontenc}   
\usepackage{lipsum}
\usepackage{color}
\usepackage{colortbl}
\usepackage{booktabs}
\usepackage{textcomp}
\usepackage{microtype}        
\usepackage{ccicons}          
\usepackage{multirow}  
\usepackage{color,soul}  
\usepackage{subcaption}  
\usepackage{listings}  
\usepackage{todonotes}
\usepackage{multirow}

\usepackage{array}

\newcommand{\thomas}[1]{\textbf{\sffamily{\textcolor{brown}{[#1 -- TP]}}}}

\usepackage{array}
\newcolumntype{P}[1]{>{\centering\arraybackslash}p{#1}}
\newcolumntype{M}[1]{>{\centering\arraybackslash}m{#1}}

\usepackage{fancyhdr}
\begin{document}

\title[Transfer Learning in Human Activity Recognition]{Transfer Learning in Human  Activity Recognition: A Survey}

\author{Sourish Gunesh Dhekane, Thomas Pl{\"o}tz}
\authornote{This work is partially supported by CISCO.}
\email{sourish.dhekane@gatech.edu, thomas.ploetz@gatech.edu}
\affiliation{%
  \institution{School of Interactive Computing, College of Computing, Georgia Institute of Technology}
  \city{Atlanta, GA}
  \country{USA}
}

\renewcommand{\shortauthors}{S.\ Dhekane and T.\ Pl{\"o}tz}

\begin{abstract}
Sensor-based human activity recognition (HAR) has been an active research area, owing to its applications in smart environments, assisted living, fitness, healthcare, etc.
Recently, deep learning based end-to-end training has resulted in state-of-the-art performance in domains such as computer vision and natural language, where large amounts of annotated data are available. 
However, large quantities of annotated data are not available for sensor-based HAR. 
Moreover, the real-world settings on which the HAR is performed differ in terms of sensor modalities, classification tasks, and target users.
To address this problem, transfer learning has been employed extensively. 
In this survey, we focus on these transfer learning methods in the application domains of smart home and wearables-based HAR.
In particular, we provide a \textit{problem-solution} perspective by categorizing and presenting the works in terms of their contributions and the challenges they address.
We also present an updated view of the state-of-the-art for both application domains. 
Based on our analysis of $205$ papers, we highlight the gaps in the literature and provide a roadmap for addressing them. 
This survey provides a reference to the HAR community, by summarizing the existing works and providing a promising research agenda.

\end{abstract}




\maketitle

 \thispagestyle{fancy}
\fancyhead{} 
\fancyfoot{} 
\pagenumbering{gobble}
\fancyfoot[C]{\textcolor{red}{This manuscript is under review. Please write to sourish.dhekane@gatech.edu for up-to-date information}}

\section{Introduction}
\label{sec:introduction}
\input{introduction}

\section{Background}
\label{sec:background}
\input{background}

\section{Transfer Learning for HAR in Smart Home Scenarios}
\label{sec:smart_home}
\input{smart_home}

\section{Transfer Learning for HAR using Wearables}
\label{sec:wearables}

\input{wearable}



\section{Discussion}
\label{sec:discussion}

\input{discussion}

\section{Conclusion}
\label{sec:conclusion}
\input{conclusion}
\bibliographystyle{ACM-Reference-Format}
\bibliography{sample-base}

\newpage

\section{Appendix}
\label{sec:appendix}
\input{appendix}

\end{document}

%% file: introduction.tex
Human activity recognition (HAR) for building intelligent systems has been a focal point of research over the last few decades, considering its applications in domains like healthcare \cite{alemdar2010wireless}, surveillance \cite{vishwakarma2013survey}, and sports \cite{adel2022survey}. 
The progress in this domain has mainly been enabled by advances in sensing technologies \cite{chen2012re}, and most importantly, machine learning \cite{lara2012survey}. 
The goal of HAR is to use the data collected by sensors placed at strategic locations (on-body or embedded in the environment) and classify it into the intended target activities. 
The workflow of HAR has been traditionally viewed as a five-step process, containing: (\textit{i}) data collection, (\textit{ii}) pre-processing, (\textit{iii}) segmentation, (\textit{iv}) feature extraction, and (\textit{v}) classification \cite{bulling2014tutorial}. 
The biggest challenge in front of the HAR community lies in the first step of this workflow, i.e., data collection \cite{plotz2023if}.

Collection of sensor data, and more importantly, annotating it correctly, is a challenging task.  
Unlike visual or language data, which can be scraped from the internet, collecting sensor data requires a sensor installation process, user consent, and manual efforts. 
Moreover, unlike images, videos, or language data instances, sensor data is not human interpretable, which makes it extremely hard to annotate the collected data. 
Thus, collecting labeled data for every specific HAR problem and training an end-to-end system to perform activity recognition is not practical.  
Transfer learning offers an alternate, and more practical, solution to this problem by leveraging the knowledge gained from performing a task on the source domain into a new task to be performed on the target domain. 
In domains like vision and language, transfer learning has been proven to be successful in achieving data efficiency, generalization, and adaptation to new tasks \cite{tan2018survey,gan2022vision,csurka2017domain,hossain2019comprehensive,zou2020survey}. 
However, performing transfer learning in the domain of HAR is not straightforward and there are substantial challenges that need to be addressed. 

The biggest technical challenge in performing transfer learning for HAR is the difference in feature space induced by factors like different sensor modalities, locations, devices, etc., across domains. 
Unlike images, videos, or text, where data instances from both domains can easily be mapped on a common feature space, such is not the case in many HAR applications. 
Thus, mapping of data points on a common feature space is a de-facto challenge to address in such applications. 
Another challenge in HAR is the unavailability of massive annotated datasets like ImageNet \cite{deng2009imagenet}, which enable researchers to design high-capacity neural networks that can generalize across settings. 
Instead, the HAR domain has multiple small-scale datasets, each with its own setting in terms of the sensor modality, device, frequency, and so on. 
Thus, addressing this limitation posed by the lack of such large-scale homogeneous labeled datasets is a prominent research interest. 
Finally, the transfer learning solutions should also be consistent with the use cases for which they are designed. 
In other words, these solutions must obey certain constraints that are imposed by the problem setting. 
For instance, the requirement of having a model well-suited for a specific set of individuals (personalization) or the requirement of data privacy (federated/distributed learning) are some of the additional challenges that need to be tackled, in addition to performing a successful knowledge transfer across domains. 
In conclusion, transfer learning in HAR is an active research area with several unsolved challenges that carry huge significance. 

Through this survey, we explored and analyzed the solutions proposed for addressing these challenges.
The specific contributions of our survey are as follows:
\begin{itemize}
    \item We provide background and define the problem of HAR from the perspectives of \textit{human activities} and \textit{sensors} so that a new researcher in this field can get introduced formally to the problem setting. 
    We also define transfer learning and provide both the problem and the solution perspectives suitable for HAR.
    \item As our main contributions, we select the application domains of smart homes and wearables to collect over 205 works that perform transfer learning. 
    We classify these works based on the challenges that they address as well as the methods they employ. 
    We present this analysis in the \textit{problem-solution} perspective, where for each challenge, its description, relevance, source and target domain settings, and approaches for addressing it are presented.
    \item After presenting this analysis, we enumerate the state-of-the-art transfer learning works from both smart home and wearable HAR domains along with their contributions in terms of the challenges addressed, novel methodologies used, comparisons performed, and performances obtained. 
    \item Based on our literature exploration and review of the state-of-the-art, we locate the gaps in the literature in terms of unaddressed challenges and unused methodologies that have the potential to perform well in addressing said challenges. 
    We then provide a roadmap for the HAR community to focus on those highlighted challenges and methods in future works.  
\end{itemize}
With this survey, we aim to direct the focus of the HAR community toward the unaddressed challenges and opportunities present in the domain of transfer learning with our stated research agenda. 
Although several works have obtained decent levels of performance in specific HAR settings, learning a generalized HAR framework that can perform inference across diverse settings, without a lot of additional adjustments is essential for the community to move forward in this field. 
We hope that this survey serves that purpose.

%% file: background.tex
\begin{figure}[]
  \centering
  \includegraphics[width=\linewidth]{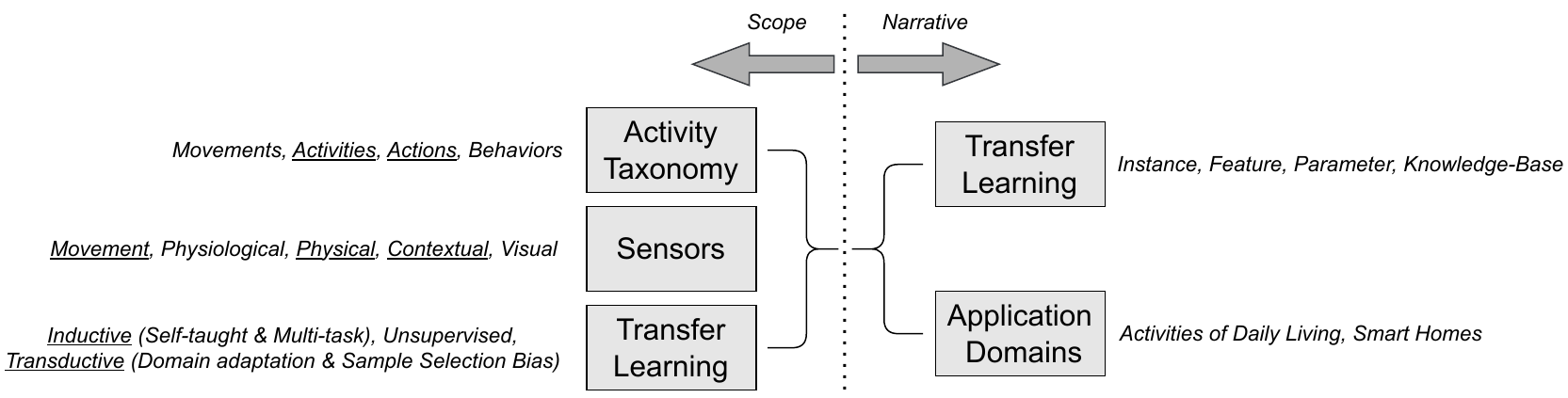}
  \caption{Concept map, where we present both the scope and the narrative of our survey. The items highlighted are included in the scope. The narrative shows the categorization of the solution space presented in the survey.
  }
  \label{fig:concept_map}
\end{figure}

This survey covers two main concepts: \textit{human activity recognition} and \textit{transfer learning}.
Before moving into their exploration, we first provide
the necessary background and introduce terminologies present in these domains.
We also use these terminologies to define the scope of our survey and present the concept map in Figure \ref{fig:concept_map}.

\subsection{Human Activities 
}
\label{subsec:HAR}

\begin{figure}[]
  \centering
  \includegraphics[width=0.75\linewidth]{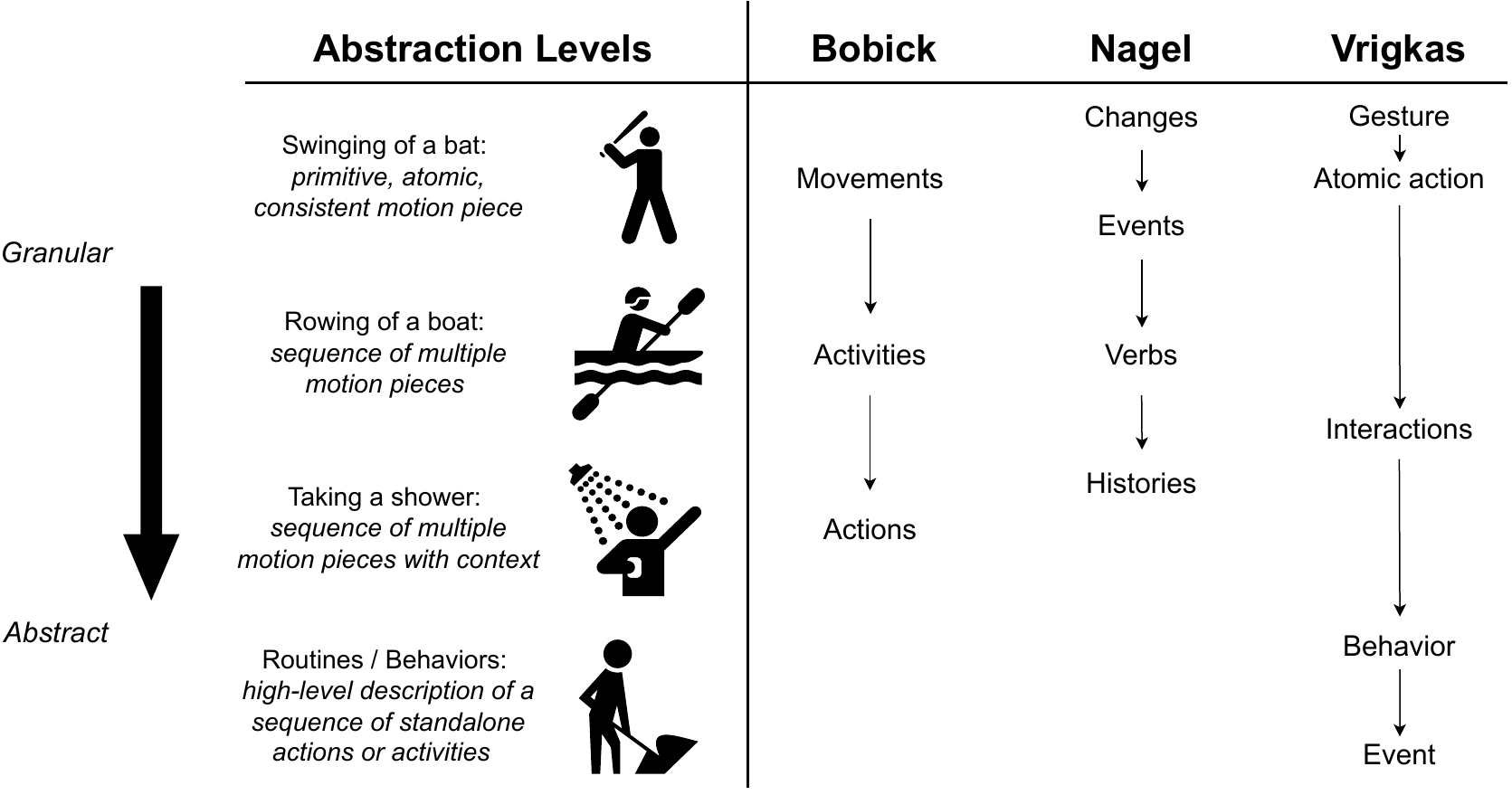}
  \caption{Illustration of the taxonomy of human activities: The interpretation of the term \textit{activity} varies across different abstraction levels. We plot the terminologies proposed by Bobick \cite{bobick1997movement}, Nagel \cite{nagel1988image}, and Vrigkas et al. \cite{vrigkas2015review} as per their level of abstraction from a granular to abstract direction. We also provide examples to visualize these terminologies. }
  \label{fig:taxonomy}
\end{figure}

Numerous works have contributed towards describing the term \textit{activity} in the context of human activity recognition \cite{vrigkas2015review,nagel1988image,bobick1997movement}.
Vrigkas et al.\ \cite{vrigkas2015review} describe \textit{activity} as a task performed by a human using physical motion and classify it into (\textit{i}) a gesture (primitive movements),
(\textit{ii}) an atomic action (single motion piece), (\textit{iii}) an interaction with the environment, (\textit{iv}) a behavior (sequence of actions representing the emotional state), or (\textit{v}) an event (high-level context-aware occurrences).
These categories cover the entire range of abstraction levels (Figure \ref{fig:taxonomy}), starting from gestures and atomic actions (granular) to behaviors and events (abstract), where activities lie in between these extremes. 
This extent of interpretation forms the basis of the taxonomy of human activities.

One of the initial
hierarchies for activities was proposed by Nagel \cite{nagel1988image}, where the terms
changes, events, verbs, and histories were used to describe deviations in signals, activity descriptors, activities, and extended sequences of activities, respectively.
Bobick's work \cite{bobick1997movement} added more description to each level of this hierarchy and presented a new system consisting of movements, activities, and actions. 
Bobick defined movements as atomic, predictable, and consistent motion units characterized by a space-time trajectory.
Based on this, activities were defined as sequences of movements with consistent patterns, and actions were defined as large-scale events with a context that lies at the intersection of perception and cognition. 
Here, actions lie on a higher level of abstraction than activities, however, a majority of literature \cite{munro2020multi, herath2017going, turaga2008machine, chaaraoui2012review} either considers actions as more granular than activities or uses these two terms interchangeably, depending on the use case.
Thus, in this survey, we consider both activities and actions as in scope (Figure \ref{fig:concept_map}).
In Figure \ref{fig:taxonomy}, we plot these previously discussed terms based on their levels of abstraction and provide examples to clarify their real-life interpretations.   


\subsection{Sensors}
The National Institute of Standards and Technology (NIST)
defines sensors as "transducers that convert a physical, biological, or chemical parameter into electrical signals" 
\cite{NIST_2021}. 
In the domain of HAR, sensors are considered as such physical devices that can collect data originating from human movements, activities, actions, and behaviors occurring in their environment.
From the perspective of data modalities, various physical quantities including motion (e.g., acceleration, angular velocities, orientation), physiological measures (e.g., heart rate, blood pressure, oxygen levels, sweating), temperature, contextual information (e.g., relative positioning, radio-frequency identifiers, open/close detection), audio, visual (e.g., RGB, depth, thermal) data can be measured \cite{chen2012re}. 
In the HAR literature, different types of sensing modalities are used in different application domains. 
Broadly, they are categorized into wearable and external sensors \cite{lara2012survey}. 
The most commonly used wearable sensors are on-body inertial measurement units (IMUs) located at different sensing positions like wrist, chest, waist, and ankle. 
An IMU typically consists of accelerometer, gyroscope, and magnetometer sensors, which collect tri-axial accelerations, angular velocities, and magnetic field readings, respectively.  
Some wearable devices are also capable of measuring physical (temperature and humidity) or physiological (ECG and blood pressure) signals. 

In the case of external sensors, they are typically installed in locations of interest in a smart environment. 
The data collected by these sensors either represents an interaction of humans with target objects like doors, kitchen appliances, and switches or a measurement of physical quantities like motion, temperature, and pressure in the vicinity of the installed sensor.  
Such sensors, also called ambient sensors, are mostly used in smart home environments to recognize activities of daily living (ADL). 
Visual sensors such as RGB, Kinect, and depth cameras are also considered among external sensors, which are primarily useful for security, surveillance, and interactive applications.     
The focus of this survey, as shown in Figure \ref{fig:concept_map}, will be on sensor-based HAR. 
In particular, we focus on sensors that measure aspects related to movements (e.g., IMU, motion sensors), physical measures (e.g., temperature, pressure sensors), and contextual measures (e.g., open/close switches) in the scope of our survey. 

\subsection{Human Activity Recognition}
Referring to the description of human activities and sensors, we provide a formal definition of HAR, inspired from the work by Lara \& Labrador \cite{lara2012survey}. 
A typical HAR setup consists of a set of sensors $S = \{S^1, S^2,...,S^n\}$ deployed for data collection, where each sensor $S^i$ collects sensor readings $\{S^i_{t_1}, S^i_{t_1+1},...,S^i_{t_2}\}$ in the time interval $T$ from timestamps $t_1$ to $t_2$. 
The problem of HAR is defined as: (\textit{i}) partitioning $T$ into intervals $\{T_1,T_2,...,T_k\}$ such that $\bigcup_{i=1}^{k} T_i = T$ and (\textit{ii}) allotting each of this $T_i = \{t_m, t_{m+1},...,t_n\}$ an activity label using the corresponding sensor data $\{S^i_{t_m}, S^i_{t_m+1},...,S^i_{t_n}\}$.
This problem is complex in nature as a large number of activities may take place in overlapping intervals. 
Thus, a relaxed version of this problem can be formulated by assuming a sliding window protocol and defining a target activity set. 
In particular, we can divide the time interval $T$ in partially-overlapping windows of equal size $\{W_1, W_2,..., W_n\}$ and assume a target activity set $Y=\{Y_1, Y_2,..., Y_l\}$ to reformulate the HAR problem as assigning each window $W_i$ with an activity label $Y_j$. 

The commonly used approach to solve this problem is via the activity recognition chain \cite{bulling2014tutorial}, which consists of five stages: (\textit{i}) data acquisition, (\textit{ii}) signal preprocessing, (\textit{iii}) segmentation, (\textit{iv}) feature extraction, and (\textit{v}) classification. 
The research works in HAR focus on improving these individual stages by targeting challenges in developing sensing technologies, setting up data acquisition systems, better feature extraction \& inference/classification frameworks, and many more areas.
One such area is improving the ability of the HAR systems to generalize in the presence of a change in sensing environments and target activities.
Transfer learning is the most intuitive way in which this challenge can be addressed.


\subsection{Transfer Learning}
\label{subsec:TLDA}

\begin{figure}[]
  \centering
  \includegraphics[width=0.95\linewidth]{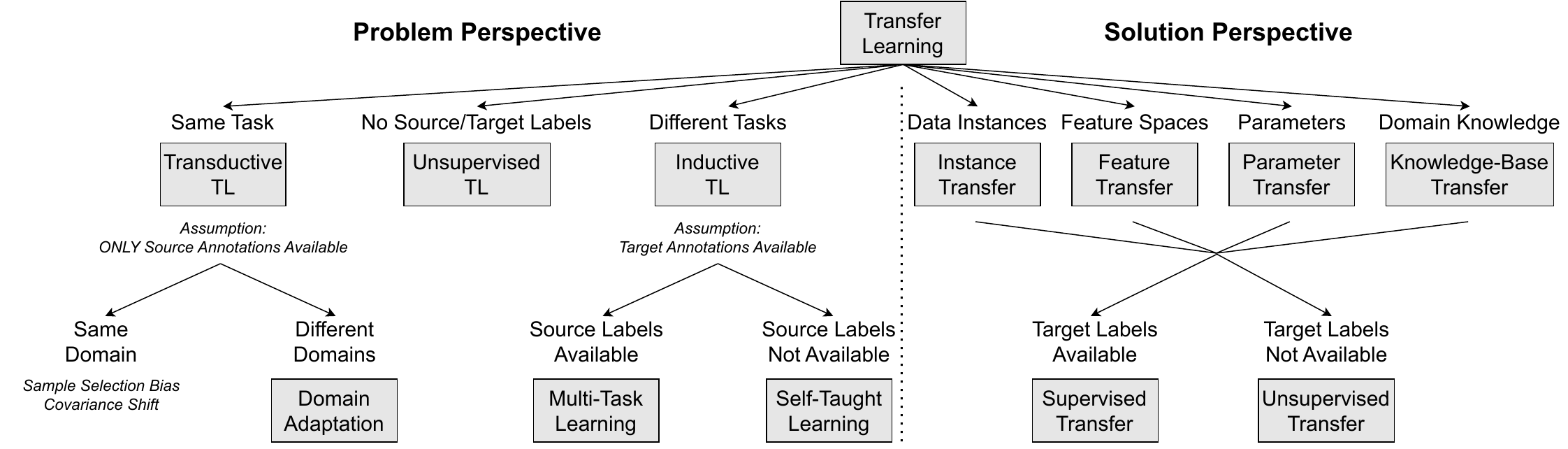}
  \caption{Categorizations in Transfer Learning inspired from the works of \cite{pan2009survey} and \cite{weiss2016survey}. Here we provide both the problem perspective, where transfer learning approaches are classified based on the characteristics of the problem setting, and the solution perspective, where the classification is based on the kind of knowledge which is transferred across the domains. 
  }
  \Description{Categorizations in Transfer Learning}
  \label{fig:TLDA}
\end{figure}

In machine learning, a supervised classification problem aims at learning a predictive function $f\colon X\to Y$, where $X=\{X_1, X_2,...,X_{m}\}$ represents the set of data points and $Y=\{Y_1, Y_2,...,Y_{k}\}$ represents the set of class labels. 
Each data point $X_i$ can be represented by an $n$ dimensional feature vector $[X^1_i, X^2_i,...,X^n_i]$ from the vector space $\chi$. 
The set of data points $X$ follows a marginal distribution $P(X)$. 
Referring to this notation, a domain $D$ is defined by the underlying feature vector space $\chi$ and the marginal distribution $P(X)$. 
A task $T$, such as supervised classification, is defined by the label set $Y$ and the predictive function $f$. 

A domain $D_1$ differs from another domain $D_2$ if their underlying feature spaces or marginal probability distributions differ, i.e.,\ $\chi_1 \neq \chi_2$ or $P(X_1) \neq P(X_2)$. 
Likewise, a task $T_1$ differs from another task $T_2$ if their respective label sets or predictive functions, in terms of conditional probabilities, differ, i.e.,\ $Y_1 \neq Y_2$ or $P(Y_1|X_1) \neq P(Y_2|X_2)$. 
On the basis of this formulation, transfer learning is defined as a set of methodologies that help improve the performance of task $T_t$ on domain $D_t$ using the \textit{knowledge} gained by performing the task $T_s$ on the domain $D_s$, where either $D_s \neq D_t$ or $T_s \neq T_t$ or both. 
In this context, $D_s$ and $D_t$ are referred to as the source and the target domains, respectively. 
This definition of transfer learning consists of two important components: (\textit{i}) the problem setting (similarity or differences in domains and tasks) and (\textit{ii}) the interpretation of the term \textit{knowledge}. 
This gives rise to the categorization of transfer learning into the problem and the solution perspectives. 

The problem perspective, as shown in Figure \ref{fig:TLDA}, classifies transfer learning based on whether the underlying domains or tasks across the source and target domains are the same. 
Specifically, transfer learning is classified into Transductive and Inductive problems, if $T_s = T_t$ and $T_s \neq T_t$, respectively. 
Since $T_s = T_t$ for transductive transfer learning, $D_s$ has to be different from $D_t$ in order for the domains to differ.
This case is referred to as Domain Adaptation. 
In the literature there also exist cases where both the domains and the tasks across settings are the same, yet there is presence of sample selection bias or covariate shift. 
Such cases are also assumed to be under the purview of transfer learning.
Inductive transfer learning is further classified into Multi-task and Self-taught Learning on the basis of the availability of the source labels, i.e.,
$\{Y_i^s\}_{i=1}^{m}$.
From this problem perspective, the domain of transfer learning mostly covers these four settings, i.e., domain adaptation, sample selection bias / covariate shift, multi-task learning, and self-taught learning, which are included in our survey's scope (Figure \ref{fig:concept_map}). 
In the remaining cases where none of the source or target labels are available, the learning problem becomes unsupervised in nature, consisting of tasks like clustering, dimensionality reduction, and density estimation.

The solution perspective of transfer learning, as shown in Figure \ref{fig:TLDA}, divides transfer learning approaches into instance, feature, parameter, and knowledge-base transfer.  
Referring to the definition of transfer learning, these types originate from whether data instances, feature spaces, model parameters, or domain knowledge are used to bolster the target domain performance.
Table \ref{tab:TL_approaches} summarizes the key themes explored under the transfer learning methods.
A brief background on the themes explored under the solutions perspective is presented in Section \ref{sec:appendix} (Appendix). 
The solution perspective is further classified into supervised and unsupervised approaches, depending on whether the target domain labels are available. 

\begin{table}
\caption{The solution perspective of transfer learning. We divide the approaches into instance, feature, parameter, and knowledge-base transfer, based on \textit{what} they transfer. This table represents the key themes explored in each category. }
\label{tab:TL_approaches}
\begin{adjustbox}{max width=\textwidth} 
  \begin{tabular}{M{6cm}M{16cm}}
    \toprule
     Types & Methods  \\
    \midrule
    Instance Transfer & Instance Weighing, Instance Mapping \& Label Propagation, Soure Domain Selection, Synthetic Data Generation \\
    Feature Space Transfer & Conventional, Neural Network Based: (a) Architectures, (b) Loss Functions, (c) Training Strategies \\
    Parameter Transfer & Conventional, Neural Network Based: (a) Architectures, (b) Loss Functions, (c) Training Strategies \\
    Knowledge Base Transfer & Domain-specific Metadata, Heuristics, Context Information \\
  \bottomrule
\end{tabular}
\end{adjustbox}
\end{table}

\subsection{Transfer Learning in Sensor-based HAR}
The literature on transfer learning is vast \cite{pan2009survey,weiss2016survey,zhuang2020comprehensive}, where domains like computer vision and natural language have been known for their extensive explorations of different methodologies. 
However, performing transfer learning for sensor-based HAR comes with its unique challenges, which require careful adaptation of the existing transfer learning methods as well as the development of the new ones. 

\subsubsection{Lack of Labeled Data}
The availability of large and diverse amounts of labeled data is crucial for achieving generalizability and transferability in machine learning. 
While large-scale data annotation projects have been undertaken in other domains \cite{deng2009imagenet}, annotating sensor data is both time and resource-consuming because of less interpretability.
Moreover, collecting even the unlabeled sensor data is difficult as it comes with logistical as well as ethical challenges including installation of sensors, recruitment of human subjects, and addressing their privacy concerns \cite{dhekane2023much}. 
Even after addressing these concerns, several factors can induce a change in the underlying distribution of the sensor data like sensor setups, location/position, sensing frequencies, orientation, user idiosyncrasies, behavior, etc. 
To this date, there are no large-scale data collection efforts in either wearable or smart home settings, covering all these sources of variations described above.   
All these challenges result in sensor-based HAR datasets being smaller in size, imbalanced in nature, and non-conducive to performing transfer learning \cite{plotz2018deep,kwon2020imutube}.

\subsubsection{Noise}
Sensor data is inherently noisy. 
This noise originates from sources like the hardware components of sensors
\cite{mohd2003measurement}, environmental conditions (autocalibration and placement of the sensor), and the way subjects perform activities \cite{plotz2023if}.  
Additionally, the other type of noise originates from the correctly sensed data that does not correspond to the target activity \cite{li2023human}. 
Since a change in domain is likely to result in a change in the noise characteristics as well as sources, performing a transfer that addresses this noise variation is a demanding task. 

\subsubsection{Learning Generalizable Features}
It has been observed that deep convolutional neural networks learn hierarchical features on visual datasets \cite{qin2018convolutional}, where shallow layers learn more domain-invariant features and deeper layers learn task-specific features. 
Such is not the case when it comes to the sensor data because of the data characteristics \cite{morales2016deep}.
Moreover, it has also been observed that the choice of the domain, in terms of the variability in the activities, plays a crucial role. 
Training performed on a more diverse activity set results in the low-level features being more generalizable \cite{morales2016deep}.
An empirical study designed to test the role of features indicated that no set of features is optimal across different HAR problems \cite{haresamudram2019role}.
This makes learning optimal features under transfer from the available domains a challenge unique to sensor-based HAR. 


\subsubsection{Sensor Heterogeneity}
In a typical transfer learning problem in vision or natural language, despite the change in domains, the data modality remains constant. 
However, this assumption is not valid, especially in the case of HAR in smart homes, where the source and the target domains may contain different numbers and types of ambient sensors. 
Even in the case of wearables, sensor heterogeneity can be present in the form of differences in sensing devices, hardware, sensing frequency, area of coverage, and positions/locations \cite{chakma2023domain}.
Tackling this heterogeneity across domains is yet another unique challenge, where the goal is to find common representation spaces that are conducive to performing a transfer. 

Addressing these challenges requires the use of specialized transfer learning methods for sensor-based HAR.
In this survey, we identify two main application domains in sensor-based HAR: (\textit{i}) Smart Home HAR and (\textit{ii}) Wearable HAR. 
The narrative of our survey focuses on highlighting the transfer learning approaches and their relevance in these two application domains, as shown in Figure \ref{fig:concept_map}.

%% file: smart_home.tex
The idea behind smart homes is to \textit{install intelligence} into the living spaces for automating day-to-day functionalities. 
In addition to automation, such intelligent spaces are beneficial for healthcare and monitoring purposes as well, especially for the elderly and at-risk population \cite{alam2012review}.
Accurate recognition of human activities plays a crucial role in these applications. 
To perform HAR in a non-obstrusive way, ambient sensors are embedded into the physical spaces, which collect data corresponding to state-changes and physical quantities like temperature, pressure, humidity, etc.
Using this data, several works \cite{bouchabou2021survey, babangida2022internet} have targeted building HAR frameworks.
Despite their promising performance, these HAR models trained on a specific smart home setting lack transferability \cite{cook2013transfer}.

Learning setting-invariant HAR models is extremely relevant in case of smart homes as they typically have different layouts, sensor installations, sensor locations, and target activities. 
Collecting large amounts of data, each time, on such specific settings and training HAR models becomes infeasible in real-life scenarios \cite{hiremath2022bootstrapping, hiremath2023lifespan}.
Thus, transfer learning for HAR in smart homes is an active research direction. 
The standard version of this problem statement of performing transfer consists of a single source $D^s$ and target $D^t$ domain, where the feature spaces are different, i.e., $\chi_s \neq \chi_t$. 
This difference arises from the variations in smart home layouts in terms of sensors, their locations, modalities, coverage, etc. 
In a simpler case, the target activities across domains remain the same, i.e., $Y^s = Y^t$ \cite{chiang2017feature, yu2023fine, lu2014instantiation}.
However, in a generic case, both the feature spaces as well as label sets vary across domains, i.e., $\chi_s \neq \chi_t$ and $Y^s \neq Y^t$ \cite{rashidi2011activity, azkune2020cross, rahman2022enabling}.
In addition to this, there are several practical challenges that are tackled in the literature including multi-source transfer, personalized transfer, optimal source selection, and more.  
In this section, we describe these challenges, their relevance, and provide details on the approaches used to address them.  

\subsection{Heterogeneous Transfer}
As mentioned, heterogeneous transfer learning refers to the case where the source and target feature spaces differ, i.e., $\chi_s \neq \chi_t$. 
There are two main interpretations of the term \textit{heterogeneous} in this context, where the change in feature space is induced by: (\textit{i}) a change in sensor modalities across the domains, or 
(\textit{ii}) a change in other aspects (e.g.,
sensor layout, cardinality, location, and sampling frequency).

There exist sensors that capture data in the discrete (e.g., ON/OFF and OPEN/CLOSE) as well as continuous (numeric) space (e.g., 21.5, 20, 19.5).  
Moreover, these sensors often belong to a functional group, where they are attached to specific objects (e.g., entrance doors) and capture information specific to them (opening and closing). 
This set of OPEN/CLOSE readings is different from another set of OPEN/CLOSE readings originating from a sensor attached to a fridge door.
The change in smart home layouts results in an uneven cardinality of sensors embedded into each meaningful location like kitchen, bathroom, and living space. 
Therefore, the number of views available for sensing an activity occurring in (say) a kitchen is often different for different domains. 
All these factors complicate the transfer learning process and require special attention to avoid performance degradation.  
Specific source-target dataset pairs used in the literature to perform heterogeneous transfer are shown in Table \ref{tab:smart_home_datasets_challenges}. 

\begin{table}
\caption{Exploration of transfer learning challenges in smart home literature with corresponding source-target dataset pairs: 
In this table, we categorize the transfer learning works in terms of the challenges they address and the source-target dataset pairs on which their experiments are performed. 
Addressing heterogeneity and task differences across domains are the most tackled challenges in the literature, whereas, CASAS and Kasteren datasets are the popular choices to perform experiments. 
In this table, the "Others" dataset category denotes custom datasets collected specific to some works as well as the lesser-used publicly available datasets 
like CASAS Single Resident, PUCK-Parkinson's, Intel, and TU Darmstadt dataset. }
\label{tab:smart_home_datasets_challenges}
\begin{adjustbox}{max width=\textwidth} 
  \begin{tabular}{M{2.5cm}M{4cm}M{4cm}M{4cm}M{4cm}M{4cm}}
    \toprule
     Datasets & Heterogeneous Transfer & Difference in Tasks & Multi-Source Transfer & Personalized Transfer  & Optimal Source Selection  \\
    \midrule
    CASAS Apartments & \cite{rashidi2011activity, rashidi2010activity, rashidi2010multi,heterogeneous_feuz_2014} & \cite{rashidi2011activity, rashidi2010activity, rashidi2010multi}& \cite{rashidi2011activity, rashidi2010activity, rashidi2010multi,heterogeneous_feuz_2014} & - & -\\ 
    CASAS HH & \cite{yu2023fine, yu2023optimal,sukhija2018label,rahman2022enabling, medrano2019enabling,niu2020multi,niu2022source} & \cite{rahman2022enabling,sukhija2018label}& \cite{niu2020multi}& - & \cite{yu2023optimal,niu2022source}\\
    CASAS (others) & \cite{feuz2017collegial, guo2023transferred,chiang2012knowledge,cook2010learning,sanabria2020unsupervised,ye2014usmart,ye2020xlearn, samarah2018transferring}   & \cite{rahman2022enabling,ye2020xlearn}& \cite{ye2020xlearn}& \cite{feuz2017collegial, rashidi2009keeping}& - \\
    MIT MAS622J & \cite{chiang2017feature,azkune2020cross,chiang2012knowledge,dridi2022transfer,ye2014usmart} & \cite{azkune2020cross}& \cite{azkune2020cross}& - & - \\
    MIT PLIA1 &  \cite{hu2011transfer} & \cite{zheng2009cross,hu2011transfer}& - & - & - \\
    Ordonez & \cite{wang2018activity, polo2020domain,van2008recognizing} & \cite{wang2018activity}& - & - & - \\
    Kasteren & \cite{ye2018slearn, sanabria2020unsupervised, sanabria2021unsupervised,ye2014usmart, ye2020xlearn, van2010transferring,azkune2020cross, hu2011transfer} & \cite{zheng2009cross,hu2011cross, azkune2020cross,ye2018slearn,ye2020xlearn,hu2011transfer}& \cite{van2010transferring,azkune2020cross,ye2018slearn,ye2020xlearn}& - & - \\
    Others & \cite{hu2011transfer, feuz2015transfer} & \cite{alam2017unseen, alam2017unseen,ramamurthy2021star,hu2011transfer}& \cite{feuz2015transfer}& \cite{diethe2015bayesian,ali2023user, ali2019improving,lu2013hybrid, rashidi2009transferring}& \cite{rashidi2011domain, sonia2020transfer}\\
  \bottomrule
\end{tabular}
\end{adjustbox}
\end{table}

\subsubsection{Approaches}
Since the source and the target feature spaces differ in heterogeneous transfer learning, a transfer of feature spaces is imperative to tackle this challenge.
There are two themes of approaches observed in the relevant works. 
In the first, a mapping function between the source and the target domain features is learned \cite{guo2023transferred,ye2018slearn,chiang2017feature,heterogeneous_feuz_2014}. 
Based on this mapping function, the source model or instances can be used in the target domain training. 
In the second theme, a transformation is applied (learned) on both the source and the target feature spaces to arrive at a common space, where models can be shared across domains \cite{feuz2017collegial,yu2023fine,niu2022source,sanabria2021unsupervised}. 
In this approach, an explicit mapping between source and target feature sets does not take place. 

The mapping process can be divided into two parts. 
First, the goal is to generate common and relevant feature representations, which is followed by the actual mapping process between these representations. 
Sensor profiling is one of the easier ways in which representations for mapping can be constructed. 
In the profiling-based approaches, the domain-specific contextual information about the sensors is used.
Thus, these approaches can also be categorized under knowledge base transfer. 
The profiling can be designed by constructing a sensor mapping/similarity matrix based on factors like the sensor type, location, attached object, trigger ratio per activity, and time information \cite{guo2023transferred, sanabria2020unsupervised}. 
Another way of representing a sensor profile is by using a binary vector, where each field denotes a property related to the object to which the sensor is attached, its location, modality, and its relation to target activities \cite{chiang2012knowledge}.
Assigning ID \cite{niu2022source} or labels based on location \cite{yu2023optimal}, modality \cite{yu2023optimal}, and functional groups \cite{van2008recognizing} to sensors and matching them to perform mapping is also explored in the literature. 

Using semantic \cite{ye2018slearn} and meta-features \cite{wang2018activity} is another way in which mapping-specific representations can be constructed. 
Meta-features can be seen as \textit{features of features} in the sense that they represent specific properties of the original data. 
For example, the average time difference between sensor activations from the kitchen is a meta-feature, which is independent of the underlying sensor modality (feature set).
Meta-features can be constructed using informed (using annotations) and uninformed (without using annotations) strategies \cite{feuz2015transfer}.
Informed meta-features usually depend on sensor-activity co-occurrence, whereas, the uninformed meta-features could be constructed using statistical measures of sensor activations like frequency and time difference.   
Meta-features can also be constructed based on clusters of instances, representing cluster-specific properties \cite{ye2020xlearn}.  
Semantic features make use of the sensor/feature/domain-specific semantic profiles to generate meaningful transferable entities. 
For instance, the number of sensors attached to kitchen doors could be a semantic feature, which is transferable as well as relevant. 
In addition to semantic features, domain semantics can also be captured by ontological approaches that use domain, sensor, and activity ontologies and semantic similarities between them \cite{ye2014usmart}.
Sometimes, meta-features are also constructed similarly to semantic features, using their functionality information \cite{van2010transferring}. 

After generating representations using the discussed techniques, the mapping process can be performed using a number of techniques. 
The mapping problem can be visualized as a graph matching problem by using a similarity metric (as weights) and features/sensors (as nodes). 
Performing similarity calculations using Jensen–Shannon divergence (JSD) and graph matching using the Gale-Shapley algorithm is a popular choice in literature \cite{chiang2017feature, chen2017activity, lu2014instantiation, chiang2012knowledge, dridi2022transfer}. 
In addition to graph matching, the expectation-maximization (EM) framework is also used to \textit{learn} the optimal mapping \cite{rashidi2011activity, rashidi2010multi}. 
Feature mapping can also be solved by performing fuzzy clustering of sensor activations followed by temporal alignment using the obtained cluster \cite{polo2020domain}. 
Another way of reformulating the mapping problem is by learning a dictionary of over-complete basis vectors via sparse coding \cite{dridi2022transfer}. 

Rather than mapping the entire feature sets, a subset of \textit{pivotal} features can also be used as a bridge between the domains. 
This pivotal feature set can be identified using random forest models trained on source and target domain data \cite{sukhija2019supervised}.
For calculating the similarity between features from source and target domains, a symmetric version of Kullback–Leibler (KL) divergence is used \cite{hu2011transfer}. 
Additionally, genetic algorithm and greedy (exhaustive) approaches for mapping are also explored in the literature \cite{heterogeneous_feuz_2014}. 

Following the second theme of approaches, where transformations are applied/learned to project the feature spaces or instances, Procrustes 
analysis is explored in the literature \cite{feuz2017collegial}. 
It finds a single orthogonal transformation that minimizes the sum-of-squares distances between all pairs of data instances across domains. 
A popular theme among such approaches is to project the feature spaces using neural networks.
Word2Vec is a good example for such approaches, where word embeddings are used as the intermediate feature spaces \cite{yu2023fine}.
Another way of using Word2Vec is to identify the common sensors across the domains and freeze their projections such that the remaining projections get aligned accordingly \cite{niu2020multi}. 
It is also observed that rather than using the sensor activations as inputs to the Word2Vec model, simply using the sensor ID can improve the quality of projected representations \cite{niu2022source}. 
Neural word embeddings can also be used to perform the same functionality as Word2Vec \cite{azkune2020cross}, which is to convert the different feature spaces into a common subspace.
After this conversion, different neural networks are used to further extract transferable features. 
Architectures including long short term memory (LSTM) \cite{yu2023fine}, autoencoders (AE) \cite{rahman2022enabling}, graph AEs \cite{medrano2019enabling} based on differential pooling, and bidirectional generative adversarial networks (Bi-GANs) \cite{sanabria2021unsupervised} are used for this purpose.

In the literature, the issue of sensor heterogeneity is also tackled under the fog computing paradigm, where the communication between individual fogs (smart homes in this case) is possible using the edge devices in the IoT system \cite{samarah2018transferring}. 
To address the sensor heterogeneity challenge, a metadata description is built for abstracting a smart home’s activities.
This description is shared with other smart homes in the fog and the target house takes into consideration this shared metadata information to perform feature mapping using meta features. 

\subsection{Difference in Tasks}
Referring to the definition of a task from Section \ref{subsec:TLDA}, a change in task represents either $Y^s \neq Y^t$ or $P(Y^s|X^s) \neq P(Y^t|X^t)$.
Although the label spaces could be different, there exists an implicit assumption that they are similar to each other semantically. 
For example, the activities \textit{sleep bed} and \textit{sleep not bed} are similar in terms of the underlying action performed by the user \cite{rashidi2011activity}. 
The similarity between activities can also be in terms of their usual location of occurrence. 
For instance, the activities \textit{preparing food} and \textit{washing dishes} both take place in the kitchen area, which links them semantically in the transfer process \cite{chiang2012knowledge}.  
The ultimate goal here is to enable the HAR system to classify unseen activities with as little supervision as possible. 
Specific source-target dataset pairs used in the literature to address task difference 
are shown in Table \ref{tab:smart_home_datasets_challenges}. 

\subsubsection{Approaches}
The problem of $Y^s \neq Y^t$ can be solved using instance and parameter transfer methods. 
A straightforward application of instance transfer techniques is via label space transfer, where a mapping between $Y^s$ and $Y^t$ is learned. 
One of the simpler ways in which label space transfer can be achieved is by using activity templates, where similar activities are grouped under one label \cite{rashidi2011activity, rashidi2010multi}.
This approach requires domain heuristics, which is why it can also be classified under knowledge-based transfer. 
Another approach using domain heuristics is the representation of human activities using semantic attributes, where the similarity between the predefined and unseen activities is calculated by creating a hierarchical cluster of activity labels generated from the available datasets \cite{alam2017unseen}.
Creating class prototypes also uses domain heuristics, where a prototype is a binary vector denoting a set of transferable entities \cite{wang2018activity}.
These prototypes can later be used in zero-shot classification settings. 
For calculating the similarity between activities, web-knowledge is also used in the literature \cite{zheng2009cross, hu2011cross}. 
Based on these similarity calculations, instance transfer and pseudo labeling can be performed. 
Another way to measure the distance between activities is to represent them using neural word embeddings and calculate the distances in the embedding space \cite{azkune2020cross}. 
In label space mapping, the use of normalized Google distance (NGD) is commonplace \cite{sukhija2018label, hu2011transfer}. 
The intuition behind using it is to have representations of instances with the same/similar labels closer to each other in the transformed feature space. 

Under the paradigm of parameter sharing, the goal is to reuse the parameters of the classifier model learned on the source. 
One way of reusing the parameters is to share classifiers across domains \cite{ye2018slearn}. 
This can be achieved by identifying non-confident predictions using least confidence, margin sampling, and entropy measures and performing feature space mapping to allow the other classifier(s) to perform inference on the non-confident samples. 
The final prediction can be obtained by selecting the most certain class label. 
There have also been works that attempt learning a generalizable transfer model, which performs ensemble learning and stacked inference to obtain a model suited for the universal set of activity labels \cite{ye2020xlearn}.
Parameter sharing can also be performed by transferring the learned weights directly across domains. 
An approach following this trend has been performed in the literature \cite{ramamurthy2021star}, where deep belief networks are used to learn features without supervision using contrastive divergence.
These features are subsequently transferred and finetuned using the available labeled data.

\subsection{Multi-source Transfer Learning}
Multi-source transfer learning is an extension of the standard transfer learning approach with multiple source domains $D^{s1}, D^{s2}, ..., D^{sk}$ and a single target domain $D^{t}$. 
Despite having access to additional training data from multiple source domains, the problem is challenging w.r.t.\ redundancy handling, the possibility of negative transfer (the case when transfer learning results in a negative impact on the target
domain performance), and weighting of domain knowledge based on their similarity with the target domain. 
In the case of smart homes, there are several datasets (e.g., CASAS \cite{rashidi2009keeping}, Kasteren \cite{van2008accurate,van2011human}) that are fully/partially annotated. 
However, they have different sensor installations, smart home layouts, and other setting-specific factors. 
Therefore, the challenge lies in learning setting-invariant knowledge from multiple domains and apply it to improve the performance over the target domain. 
Specific source-target datasets for performing multi-source transfer are shown in Table \ref{tab:smart_home_datasets_challenges}. 

\subsubsection{Approaches}
Since multi-source transfer learning with $k$ source domains has the problem of potentially having $k$ different feature spaces, feature transfer methods that do not follow the one-on-one transfer approach can be applied here. 
For example, methods like meta-features \cite{ye2020xlearn}, neural word embedding \cite{azkune2020cross}, and Word2Vec \cite{niu2020multi} can be used to construct a domain-invariant feature space for multiple source and target domains.  
Following this step, training on multiple source domains can be carried out using a classifier. 
LSTM networks are, for example,  explored for this purpose \cite{niu2020multi, azkune2020cross}.

Another prominent theme to tackle the challenge of multiple source domains is to use an ensemble \cite{rashidi2011activity, rashidi2010multi,heterogeneous_feuz_2014, ye2020xlearn}. 
The ensemble can be formed by considering individual classifiers for each domain \cite{ye2020xlearn} or sensor modality \cite{heterogeneous_feuz_2014}.
For combining the predictions of each individual classifier, a weighted majority voting scheme is explored in the literature \cite{rashidi2011activity, rashidi2010multi}.
Stacking is another way in which the predictions can be combined \cite{feuz2015transfer, ye2020xlearn}.
It uses an additional classifier that takes inputs from the predictions of individual classifiers, and in turn, predicts the output label. 
Sharing of classifiers is also a solution to having multiple source datasets \cite{ye2018slearn}.
The uncertain predictions from each source domain can make use of the classifiers trained on the other source domains to potentially remove the uncertainty. 

\subsection{Personalized Transfer Learning}
Personalization is a special kind of transfer learning problem, where neither the feature space nor the label set changes, i.e., $X^s = X^t$ and $Y^s = Y^t$. 
The shift in domain is induced by difference in the way human subjects perform activities, i.e., $P(X^s) \neq P(X^t)$
Different subjects have their own tendencies to perform activities. 
Such idiosyncrasies result in large variations in the data collected across multiple subjects for the same activity label.  
In practical scenarios, building personalized HAR systems in smart homes is always the ultimate goal. 
Specific source-target dataset pairs used in the literature for personalization are shown in Table \ref{tab:smart_home_datasets_challenges}. 

\subsubsection{Approaches}
A straightforward way of achieving personalization is by incorporating user feedback into the recognition system. 
This can be done via performing user-assisted active learning \cite{lu2013hybrid}, where data points corresponding to the most uncertain predictions are requested for annotation by the user.
The uncertainty is measured using a classifier's lack of confidence and the addition of new (unseen) feature sets.
A more sophisticated user-oriented system has also been proposed \cite{rashidi2009keeping}, where the adaptation is based on using four mechanisms: direct manipulation (using a user interface), guidance, request, and smart detection.

Although keeping the user in the loop is beneficial, overdoing so may result in an intrusive system, which is why many approaches try to implicitly achieve personalization. 
Towards this, generating synthetic data corresponding to the new user is a viable option. 
This can be achieved by conducting a user survey (one-time cost) and using simulators based on a rule-guided system to generate artificial target-subject specific data \cite{ali2023user, ali2019improving}.
The same goal is also achieved by using GANs, 
which aim at transforming source instances into target instances by changing their subject-specific characteristics \cite{soleimani2021cross}. 

Personalization can be seen as a multi-view problem, where a view corresponds to the data generated by a specific subject. 
Techniques like co-training and co-EM are used in this context \cite{feuz2017collegial}. 
Learning a user-specific HAR model in a dynamic environment has also been attempted where each iteration consists of a labeling stage by the current iteration's model followed by user-assisted active learning \cite{lu2013hybrid}.
A specialized approach for cross-subject transfer was proposed \cite{diethe2015bayesian}, where a hierarchical community-online multi-class Bayes point machine is applied on a single individual. 
The prior values of this model, originally over the community weights, are replaced by Gaussian and Gamma posteriors learned from individual-specific data.

\subsection{Optimal Source Selection}
Selection of the optimal source domain for performing transfer learning can be viewed as an approach to the multi-source transfer learning problem. 
However, we separate it out from multi-source transfer as the goal here is not to incorporate multiple domains, rather choose the most compatible domain to the given target domain. 
This approach avoids the possibility of negative transfer, where the performance over the target domain degrades because of the inclusion of a non-compatible source domain.  
To perform optimal source selection, datasets shown in Table \ref{tab:smart_home_datasets_challenges} are used in the literature. 

\subsubsection{Approaches}
The problem of optimal source selection can be reduced to  calculating the (dis)similarity between domains. 
This similarity can be measured in terms of an index, which depends on common sensor positions, types, and their cardinality \cite{yu2023optimal}.
A more comprehensive set of factors for calculating the similarity between domains can be found \cite{sonia2020transfer}, where additional factors like physical location, sampling rates, number of residents, and their routines are included. 

For measuring domain dissimilarity, a distance measure called $\mathcal{H}$ distance is proposed \cite{rashidi2011domain}, which uses classifier induced divergence as a proxy for distance between domains.
Another way of calculating domain distances is by using Word2Vec \cite{niu2022source}. 
It starts by selecting $k$ common sensors among domains based on their location/object information. 
Then, Word2Vec is applied on their sensor IDs to obtain their representations. 
This projected space is reduced using principal component analysis (PCA) and distances are calculated between all possible combinations of the reduced projections of the common sensors to obtain a global distance vector for each house. 
The domain distance is then calculated using the cosine similarity between these distance vectors.

%% file: wearable.tex
The developments in wearable sensing devices like smartphones, smartwatches, and smart clothing have made it possible to recognize human activities.
These devices are integrated with IMU sensors that can collect data corresponding to human motion in terms of tri-axial acceleration, angular velocity, and magnetic field. 
Using this data and the on-device computing power, machine learning models are deployed to perform activity inference. 
A significant amount of research in the wearable HAR field is focused on training these models and improving their performance under various settings \cite{lara2012survey,plotz2023if}. 
In real-world scenarios, these settings can vary in terms of factors like the sensor devices used, their on-body position, cardinality, the intended user(s) of the HAR system, target activities to recognize, compute platforms, etc. 

The need for transfer learning in these wearable HAR systems arises from such variability in settings. 
For example, a machine learning model trained on a sports activity dataset collected from a group of children using an ankle-worn IMU sensor cannot be directly reused to infer office activities performed by adults using a smartwatch.
The fact that data collection and annotation for every such setting to train the activity recognition models from scratch is infeasible makes it even more compelling to explore transfer learning for achieving generalizability. 
The transfer learning literature in wearable HAR focuses on the challenges presented by one or more of these factors of variability. 

A change in the sensor device, number, position, sensing frequency, or modality introduces heterogeneity in the transfer learning problem.
With the availability of multiple annotated source settings, the challenge of leveraging them without consequently degrading the HAR performance also becomes relevant.  
In some cases, the target setting may dynamically change, so that the underlying HAR framework needs to be continually updated to keep up with such changing environments.
Moreover, instead of an on-device system, there may be a requirement for a distributed or federated setup. 
All these challenges are introduced by a change in the sensing platform. 
In terms of the activities to recognize, the source and target settings can vary in terms of their HAR task. 
Also, regarding the users of the HAR system, there may exist a requirement for the system to perform well for a specific set of users, which calls for personalized HAR systems. 
Finally, some works also focus on measuring the performance of the HAR systems under a transfer learning setup. 
In this section, we describe these challenges, their relevance, and provide details on the approaches used to address them.   

\subsection{Heterogeneous Transfer}
The interpretation of the term \textit{heterogeneous} in the context of transfer learning is based on the feature spaces across the domains being different, i.e., $\chi_s \neq \chi_t$.
In the case of wearable sensors, this difference arises from factors like sensor modality, cardinality, sensor position, device types, sensing frequency, etc.
Based on these factors, we classify this heterogeneity into three types: (\textit{i}) \textit{m}-heterogeneity (due to a change in sensor modality), (\textit{ii}) \textit{p}-heterogeneity (due to a change in sensor position), and (\textit{iii}) \textit{d}-heterogeneity (due to a change in device aspects).  
The \textit{m}-heterogeneity in wearable HAR typically occurs when other data modalities like RGB videos, motion capture, and web data (text) are used to bolster the IMU-based transfer. 
The intention behind this added heterogeneity is to capitalize on the large datasets or pre-trained models available in related domains to learn the motion or activity-specific information that is \textit{not} specific to the IMU sensor modality.  
In contrast to this, the \textit{p} and \textit{d}-heterogeneities are inherent to the transfer learning problems, and hence, must be addressed. 
Out of these two, the \textit{p}-heterogeneity has been observed to be more challenging as different kinds of movement patterns are recorded when the sensors are attached to different body parts. 
To address these heterogeneities, several works have used source-target pairs based on the datasets shown in Table \ref{tab:wearable_datasets_challenges_1}. 
As many works target addressing more than one heterogeneities, we categorize the works based on their focus of evaluation. 
Also, all the works deal with the \textit{d}-heterogeneity, hence, we catalog only those works under it that do not address the \textit{p} or \textit{m}-heterogeneities.

\subsubsection{Approaches for addressing \textit{m}-heterogeneity}
The introduction of \textit{m}-heterogeneity into the transfer learning is done by using datasets or pre-trained models from other modalities.
The goal here is to reuse the modality-independent information from domains where either the datasets are massive or other crucial information is present. 
In the literature, other modalities like human pose \cite{activity_okita_2018,mars_pei_2021,videobased_awasthi_2022}, images \cite{physical_ataseven_2022}, videos \cite{kwon2020imutube,activity_okita_2018,opportunistic_banos_2021}, contextual data (Wi-Fi, bluetooth, GPS) \cite{less_hu_2016}, text \cite{leng2023generating,instilling_wei_2016} are leveraged along with the IMU-based recognition. 

Human pose information is used by placing virtual sensors on the skinned multi-person linear (SMPL) mesh surface and using forward kinematic and finite differences techniques to generate virtual IMU data \cite{deep_xiao_2020}. 
This virtual data, along with the real IMU data, can be used for training the target HAR system. 
Videos are also a useful data modality considering the dataset sizes, availability of pre-trained models, and the ability to generate virtual IMU data from it. 
In terms of the pre-trained models specifically, they can be used to make predictions on video streams, thereby complementing the sensor-stream decision along with it \cite{vision_radu_2019}. 
Without using pre-trained models, video data can be used along with sensor data by learning a multi-input-multi-output mapping between data sources based on QR factorization \cite{opportunistic_banos_2021}. 
However, when both video and IMU data streams are not available simultaneously, a more suitable approach is to generate virtual IMU data \cite{kwon2020imutube}.
Some methods also use the human pose and video modalities together by converting the spatio-temporal data into discrete word representation followed by LSTM-based projection onto a common feature space \cite{activity_okita_2018}. 

\begin{table}
\caption{Exploration of transfer learning challenges of sensor heterogeneity and personalization in wearable HAR literature with corresponding source-target dataset pairs: 
In this table, we categorize the transfer learning works in terms of the challenges they address and the source-target dataset pairs on which their experiments are performed. 
The challenges of \textit{p}-heterogeneity and personalization are largely addressed in the literature, whereas, PAMAP2, OPPORTUNITY, and DSADS datasets are the popular choices to perform experiments.
In this table, the "Others" dataset category denotes custom datasets collected specific to some works as well as the lesser-used publicly available datasets like Cooking, DAS, MHAD, UFO, DIP, SAD, MobiFall, RealDisp, Oulu, JSI-FOS, Sussex-Huawei, CogAge, Capture-24, Wetlab, Daphnet, HASC, CASAS MAR/YA, MEx, iPodTouch, ExtraSensory, SHL, uWave, DUMD, ANSAMO, TRA, UCI-HAPT, and MoST. 
}
\label{tab:wearable_datasets_challenges_1}
\begin{adjustbox}{max width=\textwidth} 
  \begin{tabular}{M{2.5cm}M{5.5cm}M{5.5cm}M{5.5cm}M{5.5cm}}
    \toprule
     Datasets & \textit{m}-Heterogeneity & \textit{p}-Heterogeneity & \textit{d}-Heterogeneity & Personalization  \\
    \midrule
    HHAR & - & \cite{cnn_chikhaoui_2018,systematic_chang_2020,crossdomain_thukral_2023,deeptranshhar_kumar_2022,multitask_saeed_2019,scaling_khan_2018,strangan_faridee_2021,transferring_akbari_2019,haresamudram2022assessing,haresamudram2023investigating} & \cite{framework_wen_2017,signallevel_saeedi_2020,combining_presotto_2023} & \cite{cnn_chikhaoui_2018,framework_wen_2017,signallevel_saeedi_2020,augtoact_faridee_2019,domain_hussein_2022,learning_wen_2017,strangan_faridee_2021,towards_li_2022}\\
    PAMAP2 & \cite{alinia2020actilabel, modelagnostic_alinia_2023,learning_xia_2021,physical_ataseven_2022,videobased_awasthi_2022,kwon2020imutube,leng2023generating} & \cite{haresamudram2023investigating,haresamudram2022assessing, alinia2020actilabel, modelagnostic_alinia_2023,chakma2021activity,systematic_chang_2020,contrasgan_sanabria_2021,crossdomain_lu_2021,crossdomain_thukral_2023,crossposition_chen_2019,deep_wang_2018,deep_yan_2022,digging_hoelzemann_2020,dmstl_zhu_2023,endtoend_ye_2021,human_barbosa_2018,human_varshney_2021,learning_xia_2021,local_zhao_2021,multisource_dong_2023,recurrenthar_kumar_2022,semanticdiscriminative_lu_2022,semisupervised_chakma_2022,sensehar_jeyakumar_2019,strangan_faridee_2021,stratified_wang_2018,transferring_akbari_2019,unsupervised_barbosa_2018,deep_kondo_2022} & \cite{achieving_lago_2020,combining_presotto_2023,crossdataset_gjoreski_2019} & \cite{chakma2021activity,adversarial_suh_2022,domain_qin_2022,endtoend_ye_2021,local_zhao_2021,personalized_fallahzadeh_2023,personalized_reiss_2013,personalizing_akbari_2020,semanticdiscriminative_lu_2022,semisupervised_chakma_2022,strangan_faridee_2021,stratified_wang_2018,unsupervised_barbosa_2018,generic_soleimani_2022}\\
    OPPORTUNITY & \cite{morales2016deep,deep_li_2020,digging_hoelzemann_2020,learning_xia_2021,videobased_awasthi_2022,kwon2020imutube} & \cite{chakma2021activity,systematic_chang_2020,collaborative_kong_2020,crossposition_chen_2019,morales2016deep,deep_wang_2018,endtoend_ye_2021,learning_xia_2021,semisupervised_chakma_2022,sensehar_jeyakumar_2019,sharenlearn_rokni_2019,strangan_faridee_2021,stratified_wang_2018,synchronous_rokni_2017,transfer_du_2019,transnet_rokni_2020,additional_lago_2021} & \cite{achieving_lago_2020,crossdataset_gjoreski_2019} & \cite{closedloop_saeedi_2017,chakma2021activity,adversarial_suh_2022,endtoend_ye_2021,latent_qian_2021,semisupervised_chakma_2022,strangan_faridee_2021,stratified_wang_2018,swladapt_hu_2023,transnet_rokni_2020,generic_soleimani_2022}\\
    Mobiact & - & \cite{cnn_chikhaoui_2018,crossdomain_thukral_2023,multitask_saeed_2019,haresamudram2022assessing,haresamudram2023investigating} & \cite{combining_presotto_2023} & \cite{cnn_chikhaoui_2018,semisupervised_presotto_2022} \\
    RealWorld & \cite{learning_xia_2021,kwon2020imutube,leng2023generating} & \cite{cnn_chikhaoui_2018,systematic_chang_2020,crossdomain_thukral_2023,learning_xia_2021} & \cite{combining_presotto_2023} & \cite{cnn_chikhaoui_2018,swladapt_hu_2023}\\
    DSADS & - & \cite{chakma2021activity,contrasgan_sanabria_2021,crossdomain_lu_2021,crossposition_chen_2019,deep_wang_2018,dmstl_zhu_2023,endtoend_ye_2021,local_zhao_2021,multisource_dong_2023,semanticdiscriminative_lu_2022,semisupervised_chakma_2022,stratified_wang_2018,deep_kondo_2022} & - & \cite{chakma2021activity,domain_qin_2022,endtoend_ye_2021, local_zhao_2021,personalization_fallahzadeh_2017,personalized_fallahzadeh_2023,personalized_saeedi_2018,semanticdiscriminative_lu_2022,semisupervised_chakma_2022,stratified_wang_2018} \\
    Myogym & - & \cite{crossdomain_thukral_2023,haresamudram2022assessing,haresamudram2023investigating} & - & -\\
    MHEALTH & - & \cite{haresamudram2023investigating,haresamudram2022assessing, crossdomain_thukral_2023,crossposition_chen_2019,deep_yan_2022,human_varshney_2021,deep_kondo_2022} & - & \cite{adversarial_suh_2022}\\
    WISDM & - & \cite{contrasgan_sanabria_2021,deeptranshar_kumar_2023,multitask_saeed_2019,recurrenthar_kumar_2022,transnet_rokni_2020,deep_kondo_2022} & \cite{evaluation_islam_2019} & \cite{incremental_mazankiewicz_2020,personalized_rokni_2018,towards_li_2022,transfer_an_2020,transnet_rokni_2020}\\
    UCI & \cite{deep_hashim_2022} & \cite{crossdomain_lu_2021,feature_chen_2021,multitask_saeed_2019,scaling_khan_2018,semanticdiscriminative_lu_2022,deep_kondo_2022} & \cite{framework_wen_2017,combining_presotto_2023,domain_hussein_2022} & \cite{active_diethe_2016,framework_wen_2017,crossperson_deng_2014,latent_qian_2021,learning_wen_2017,semanticdiscriminative_lu_2022,semisupervised_hashemian_2021,transfer_an_2020}\\
    SmartSock & \cite{alinia2020actilabel, modelagnostic_alinia_2023} & \cite{alinia2020actilabel, modelagnostic_alinia_2023} & - & \cite{personalized_fallahzadeh_2023}\\
    SDA & - & \cite{collaborative_kong_2020,sharenlearn_rokni_2019,synchronous_rokni_2017,transnet_rokni_2020} & - & \cite{personalized_rokni_2018,transnet_rokni_2020} \\
    MotionSense & - & \cite{crossdomain_thukral_2023,multitask_saeed_2019,haresamudram2022assessing,haresamudram2023investigating} & \cite{combining_presotto_2023} & -\\
    SKODA & \cite{deep_hashim_2022, morales2016deep,digging_hoelzemann_2020} & \cite{morales2016deep,digging_hoelzemann_2020,transfer_du_2019} & \cite{crossdataset_gjoreski_2019} & -\\
    USC-HAD & \cite{leng2023generating,deep_hashim_2022} & \cite{haresamudram2022assessing,crossdomain_lu_2021,semanticdiscriminative_lu_2022,deep_kondo_2022} & - & \cite{active_diethe_2016,domain_qin_2022,semanticdiscriminative_lu_2022}\\
    AMASS & \cite{deep_xiao_2020, learning_xia_2021,mars_pei_2021} & \cite{learning_xia_2021} & - & -\\
    UniMiB & - & \cite{multitask_saeed_2019,deep_kondo_2022} & - & \cite{latent_qian_2021,transfer_an_2020}\\
    KU-HAR & - & \cite{deeptranshar_kumar_2023,deeptranshhar_kumar_2022,recurrenthar_kumar_2022} & \cite{transfer_pavliuk_2023} & -\\
    Others & \cite{deep_li_2020, mars_pei_2021, activity_okita_2018, instilling_wei_2016,less_hu_2016,opportunistic_banos_2021,vma_hu_2022,alinia2020actilabel, modelagnostic_alinia_2023} & \cite{transferring_akbari_2019,sharenlearn_rokni_2019,human_barbosa_2018,unsupervised_barbosa_2018,dmstl_zhu_2023,deep_kondo_2022,haresamudram2022assessing,haresamudram2023investigating,crosslocation_janko_2019, collaborative_kong_2020,feature_chen_2021, autonomous_rokni_2016,deep_yan_2022,domain_alajaji_2023,kernel_wang_2017,plugnlearn_rokni_2016,scaling_khan_2018,sensehar_jeyakumar_2019,synchronous_rokni_2017,transnet_rokni_2020,vma_hu_2022,alinia2020actilabel, modelagnostic_alinia_2023} & \cite{signallevel_saeedi_2020,crossmobile_zhao_2010,crosspeople_zhao_2011,achieving_lago_2020,framework_wen_2017,crossdataset_gjoreski_2019,evaluation_islam_2019,domain_hussein_2022,transfer_pavliuk_2023} & \cite{personalizing_akbari_2020,signallevel_saeedi_2020,adapting_gong_2022, metasense_gong_2019,augtoact_faridee_2019,crosspeople_zhao_2011,enabling_lane_2011,improving_longstaff_2010,nuactiv_cheng_2013,online_sztyler_2017,personalized_fu_2021,toward_hong_2016,transfer_li_2022,transnet_rokni_2020,generic_soleimani_2022,framework_wen_2017,domain_wilson_2022,evaluating_wijekoon_2020,importanceweighted_hachiya_2012,transfer_shen_2023,domain_hussein_2022,towards_li_2022,unsupervised_barbosa_2018,transfer_an_2020}\\
  \bottomrule
\end{tabular}
\end{adjustbox}
\end{table}

In addition to videos, the image modality has also been used, where pre-trained models on ImageNet are transferred for IMU-based recognition \cite{physical_ataseven_2022,deep_hashim_2022}. 
This is done by finetuning them on IMU data using continuous wavelet transforms such as horizontal concatenation, acceleration magnitude, and pixel-wise axes-averaging  \cite{physical_ataseven_2022}. 
Another popular choice for modality is the text data, which is used as an additional data source as well as a mode to generate virtual IMU data. 
When used as a data source, it needs to be projected on the same feature space as that of the sensor data, which is achieved using matrix factorization with maximizing the empirical likelihood \cite{instilling_wei_2016}.
For virtual IMU data generation, recent works have used large language models to generate prompts that are fed to a CLIP based model to convert them to 3D human motion sequences \cite{leng2023generating}.
These motion sequences can be converted to virtual IMU data using inverse kinematics.  
Along with these popular data modalities, where large datasets and pre-trained models are readily available, contextual data originating from WiFi, bluetooth, and GPS is also used in the literature. 
A self-supervised incremental learning approach is explored to use this data that uses adaptive resonance theory based feature learning \cite{less_hu_2016}. 
Within the wearable modality itself, \textit{m}-heterogeneity can arise when additional data sources like eye-movement and strength measurements are also tracked in addition to the IMU components. 
To address this, pre-training based on sensor modality classification is performed \cite{deep_li_2020}.

\subsubsection{Approaches for addressing \textit{p}-heterogeneity}
To address the \textit{p}-heterogeneity, several instance, feature, and parameter transfer methods have been proposed.
Under the instance transfer methods, instance mapping, instance weighing, label propagation, data augmentation, and source domain selection are the commonly explored themes. 

Instance mapping refers to a method that learns a mapping function between source and target data instances or their substructures (clusters or subsequences). 
In the literature, mapping of clusters has been preferred, which is performed based on maximum mean discrepancy (MMD) \cite{local_zhao_2021}, Wasserstein distance \cite{deeptranshar_kumar_2023, deeptranshhar_kumar_2022}, and optimal transport methods \cite{crossdomain_lu_2021}.
The general idea behind instance weighing is to assign importance to source data instances, in the form of weights, to denote their usefulness in the transfer process.
One way in which the weighing process can be performed is based on the closeness of the data instances. 
Methods like kernel mean matching (KMM) \cite{human_barbosa_2018,unsupervised_barbosa_2018}, nearest neighbor weighing \cite{human_barbosa_2018,unsupervised_barbosa_2018}, and MMD-based weighing \cite{stratified_wang_2018} follow this approach.  
Weighing can also be performed based on norms like $L_{2,1}$ matrix norm \cite{feature_chen_2021} and loss functions like domain discrimination loss \cite{endtoend_ye_2021}. 
The $L_{2,1}$ matrix norm is a sequential application of the $l_2$ and $l_1$ norms, which acts as a regularizer, whereas, the domain discrimination loss depends on the inference of a domain classifier in terms of its ability to distinguish between domains (where higher confusion represents more similar data points).   

In label propagation, the aim is to learn a mapping between source and target label sets using target and source data instances. 
A well-established way of mapping labels is to cluster the source and target instances and perform graph matching \cite{plugnlearn_rokni_2016}.
In addition to clusters, their properties like centroid distances for each data instance are also used in the literature, followed by affinity graph based propagation of labels \cite{collaborative_kong_2020}. 
Another similarity graph based label propagation technique constructs the graph by assuming data instances as nodes and the Gaussian kernel similarity between them as edge weights \cite{synchronous_rokni_2017}.
After constructing this graph, an iterative label refinement is performed, where in each iteration, the nodes take a fraction of label information from their neighborhood and maintain a distribution until it stabilizes.  

As seen earlier, instance weighting assigns importance scores (weights) to individual data instances. 
The source domain selection approach follows a similar suit, but for domains, in order to choose compatible domains for performing transfer in a multi-source setting. 
Several distance measures like stratified distance \cite{crossposition_chen_2019}, global \& subdomain distances \cite{dmstl_zhu_2023} and $\mathcal{A}$ distance, semantic distance, \& cosine similarity based kinetic distances \cite{deep_wang_2018} are proposed to measure the domain similarity. 
The goal of using data augmentations in instance transfer is to generate target-like data instances by applying transformations.  
In addition to the basic data transformations \cite{systematic_chang_2020}, the literature also consists of works that aim at attribute-specific and semantically aware augmentation techniques to follow domain constraints \cite{semanticdiscriminative_lu_2022}.  
Class augmentations is another such method that applies transformations on the data instances using metadata to generate gender and sensor position specific target instances \cite{deep_kondo_2022}.

Since \textit{p}-heterogeneity causes the feature spaces to be different, feature transfer techniques are largely explored. 
The conventional feature transfer approaches for addressing \textit{p}-heterogeneity include transfer component analysis (TCA) \cite{human_barbosa_2018,unsupervised_barbosa_2018}, subspace alignment (SA) \cite{human_barbosa_2018,unsupervised_barbosa_2018}, hierarchical clustering \cite{additional_lago_2021}, and transferrable feature selection \cite{crosslocation_janko_2019}.   
TCA uses the MMD measure to learn transfer components in a reproducing kernel Hilbert space.
In SA, the subspaces formed by the top-$k$ eigenvectors of source and target domains are used in learning the mapping between them.
Hierarchical clustering is a feature agglomeration technique that recursively merges similar features together in clusters, minimizing the variance within each cluster for common feature space learning.
In the transferrable feature selection method \cite{crosslocation_janko_2019}, the mutual information between the features and class labels is calculated to rank the features followed by the removal of redundant features based on the Pearson correlation coefficient. 
On the remaining features, a greedy feature selection method is employed that incrementally adds the high-ranked features until validation performance starts degrading. 

In addition to these conventional approaches, neural networks are also used to project source and target data instances onto a common feature space. 
To guide the training of these neural networks, loss functions like adversarial domain discrimination/confusion loss \cite{chakma2021activity,systematic_chang_2020,dmstl_zhu_2023}, contrastive loss \cite{contrasgan_sanabria_2021}, MMD loss \cite{crossposition_chen_2019,deep_wang_2018,dmstl_zhu_2023,feature_chen_2021,multisource_dong_2023}, triplet loss \cite{domain_alajaji_2023,endtoend_ye_2021}, $L_{2,1}$ matrix norm loss \cite{feature_chen_2021}, KL divergence loss \cite{scaling_khan_2018}, and large margin loss \cite{semanticdiscriminative_lu_2022} are used in the literature.  
The domain discrimination/confusion loss is calculated by performing adversarial training, where the model tries to classify the domain of the input data representations.
If the model fails to correctly classify the domain, it acts as in indicator that the resultant features are domain agnostic. 
Contrastive loss is calculated on a batch of inputs, where the projections of the positive pairs (augmentations of the same data instance) are brought closer and vice versa. 
In triplet loss, a triplet of an anchor, positive, and negative sample is considered. 
The projection of the positive from the anchor is then brought closer and the negative samples are projected farther. 
While the other loss functions rely on only the correct predictions, the large margin loss takes into account the confidence in the predictions as well, where less confident predictions are penalized.

Under the parameter transfer methods, the commonly used approach is to perform pre-training on the source domain and then finetuning the learned model on the target domain. 
Several neural network architectures like convolutional neural networks (CNN) \cite{cnn_chikhaoui_2018,domain_alajaji_2023,human_varshney_2021,scaling_khan_2018,sensehar_jeyakumar_2019,transnet_rokni_2020}, gated recurrent units (GRU) \cite{deeptranshar_kumar_2023,recurrenthar_kumar_2022}, CNN-LSTM \cite{deep_wang_2018}, CNN-GRU \cite{deeptranshhar_kumar_2022}, variational autoencoders (VAE) \cite{transferring_akbari_2019}, Bi-GAN \cite{contrasgan_sanabria_2021}, graph neural networks (GNN) \cite{deep_yan_2022}, and attention units (transformers) \cite{dmstl_zhu_2023,domain_alajaji_2023,strangan_faridee_2021} are used in the model architecture.  
Here, predefined layers of the neural network (typically the lower layers) are transferred over to the target domain. 
This transfer can take place in a cascaded manner, where trainable layers are added and transferred one by one \cite{transfer_du_2019}. 

Ensemble learning in parameter transfer uses individual base models trained on source domain(s) towards target inference.
There are several ways in which the decisions of individual models can be combined. 
Several methods like learning a gating function \cite{autonomous_rokni_2016, sharenlearn_rokni_2019}, majority voting \cite{crossposition_chen_2019,stratified_wang_2018}, and discounting proportional conflict redistribution (PCR5) rule \cite{multisource_dong_2023} are proposed in the literature to combine the individual decisions. 
Another parameter transfer approach is knowledge distillation, where a smaller student model learns to mimic a larger teacher model. 
This training process can be guided using loss functions like hard cross entropy and soft knowledge distillation loss, which is the difference between the logits of the source and the target models \cite{collaborative_kong_2020}. 
This teacher-student framework is also combined with a self-training paradigm \cite{crossdomain_thukral_2023}. 
The teacher model training on the source domain is performed in a supervised manner, which is used on the target domain to generate soft pseudo labels.
The student model is then trained using the labeled source and pseudo-labeled target data, along with a self-supervised loss.
Self-training is also performed using extreme learning machines (ELM) as the base classifier, where high-confidence pseudo labels are used in an iterative manner.   

Recently, self-supervised learning has proven to be an effective way of performing pre-training.
It uses the unlabeled data itself to generate a supervisory signal on which the models are pre-trained. 
The pre-training takes place on the source domain, whereas, the finetuning takes place on the target domain. 
Pretext tasks like multitask learning, masked reconstruction, contrastive predictive coding (CPC), enhanced CPC, autoencoders (AE), SimCLR, SimSiam, and BYOL are explored in the literature in the transfer learning setup \cite{multitask_saeed_2019, haresamudram2022assessing, haresamudram2023investigating}.    

\subsubsection{Approaches for addressing \textit{d}-heterogeneity}
Since \textit{d}-heterogeneity does not assume different sensing modalities or positions across domains, the source and target feature spaces are more similar to each other than in the case of \textit{m} or \textit{p}-heterogeneity. 
Thus, instance and parameter transfer techniques are preferred over feature transfer to deal with \textit{d}-heterogeneity. 
Only a handful of approaches like feature clustering \cite{achieving_lago_2020} and MMD loss based intermediate feature space construction \cite{domain_hussein_2022,evaluation_islam_2019} are used in the relevant literature 

Boosting is one of the common ways in which data instances can be weighted. 
Methods like AdaBoost \cite{framework_wen_2017} maintain a distribution of weights over the data instances, which gets updated based on the performance (misclassification) of the weak learner.
Instance mapping in the format of mapping subsequences using brute force, clustering-based, and motif-based methods is also explored in the literature \cite{closedloop_saeedi_2017}. 
It uses normalized cross-correlation (NCC) as the underlying similarity measure. 

Among the parameter transfer methods, the approach of \textit{pretrain-then-finetune} is greatly applicable in this scenario as the underlying feature spaces are not drastically different. 
On IMU datasets, pre-training using well-established architectures like Inception-ResNet, DeepConvLSTM, HART, and MobileHART architectures and finetuning in leave one dataset out manner has been performed \cite{combining_presotto_2023}. 
In addition, other custom architectures like spectro-temporal ResNet \cite{crossdataset_gjoreski_2019}, VGG \cite{deep_kondo_2022}, and customized CNNs \cite{evaluation_islam_2019,transfer_pavliuk_2023} are also used. 
A slight variation to this approach by replacing the finetuning step with a clustering-based inference has been tested in the literature. 
In particular, classifiers like ELMs \cite{crossmobile_zhao_2010} and decision trees \cite{crosspeople_zhao_2011} are trained on the source domain and used to perform initial inference on the target domain. 
The most confident predictions are then used as initial cluster centers for performing K-means clustering followed by label propagation until convergence.
A typical parameter transfer framework assumes a direct sharing of learned weights across domains. 
However, soft parameter sharing assumes a non-linear relation between the transferred parameters, which is suitable for domain shifts \cite{domain_hussein_2022}.

\subsubsection{Approaches for addressing \textit{p} and \textit{m}-heterogeneity}
Out of the entire collection, only a handful of approaches target and perform evaluations for addressing both \textit{p} and \textit{m}-heterogeneities.
ActiLabel is one such approach, where the transfer learning problem is formulated as a combinatorial optimization problem \cite{alinia2020actilabel, modelagnostic_alinia_2023} to tackle both heterogeneities.  
It aims at learning a mapping from the groups of similar target instances, called core clusters, to known activity classes from the source domain.
Using it, a dependency graph is created on which a min-cost mapping is performed for source-to-target inference.
In another approach, the challenge of \textit{p} and \textit{m}-heterogeneity is addressed using disentangled representation learning between different modalities and positions using mutual information based regularization \cite{learning_xia_2021}.
Moreover, an additional data source of motion capture is also leveraged using an SMPL model to generate virtual IMU data. 
Instead of addressing both the \textit{p} and \textit{m}-heterogeneity at the same time, it can also be addressed one at a time using a multi-factor transfer learning approach \cite{vma_hu_2022}. 
In this approach, the multi-factor domain variance is decoupled in a transfer path based on the available domains, i.e., rather than transferring from $S\to T$, a transfer path $S\to D_1\to D_2\to...\to T$ is constructed. 
This is achieved by selecting the modalities that are more robust for each intermediate transfer. 
To perform the intermediate transfer, a parameter transfer approach is explored, which uses a multi-task model with modality-specific input and output layers to perform inference on the intermediate domain data.
The high-confidence predictions are then used for pseudo-labeling the intermediate domain, to eventually re-train the multi-task model.   

In addition to proposing such new frameworks, the literature also consists of works that perform an analysis of existing approaches for addressing the \textit{p} and \textit{m}-heterogeneity. 
Specifically, the analysis of transferring different layers of the CNN \cite{morales2016deep} and Deep ConvLSTM \cite{digging_hoelzemann_2020} networks in terms of their freezing, re-training, and re-initialization has been performed.

\subsection{Personalized Transfer Learning}
One goal of performing transfer learning is to achieve generalization. 
It refers to the ability of a model to perform well in diverse settings. 
However, in real-world deployments, personalization is a key requirement. 
Personalization can be viewed as a challenge of building a HAR system targeted for a single user (e.g.\ users of a smartwatch or a smartphone) or a set of users with common properties (e.g.\ adults in a gym). 
Specifically, it targets the the scenario of $P(X_s) \neq P(X_t)$ while $\chi_s$ may (not) be equal to $\chi_t$. 
This difference in the marginal probability distributions arises because of the variations in the activities performed by different users, owing to their idiosyncrasies. 
To address this issue, several works have used source-target pairs based on the datasets shown in Table \ref{tab:wearable_datasets_challenges_1}.

\subsubsection{Approaches}
Before discussing the approaches, we note that many of the methods presented below address the issue of $P(X_s) \neq P(X_t)$ in addition to $\chi_s \neq \chi_t$. 
This is because the issues of personalization and heterogeneity are typically present simultaneously in real-world settings. 
Thus, many of the methods used for addressing feature space heterogeneity are also used for personalization purposes.

Under instance transfer, such common approaches include: (\textit{i}) instance mapping and label propagation of sensor subsequences \cite{closedloop_saeedi_2017} or clusters \cite{local_zhao_2021} using NCC and MMD based similarity measures, respectively, (\textit{ii}) label propagation on clusters using graph matching algorithms \cite{personalized_fallahzadeh_2023}, (\textit{iii}) standard \cite{augtoact_faridee_2019} and semantically-aware \cite{semanticdiscriminative_lu_2022} data augmentations, (\textit{iv}) synthetic data generation in adversarial setup \cite{generic_soleimani_2022}, and (\textit{v}) instance weighing based on KMM \cite{transfer_li_2022,unsupervised_barbosa_2018}, nearest neighbor weighing \cite{unsupervised_barbosa_2018}, and domain discrimination loss \cite{endtoend_ye_2021,swladapt_hu_2023}.  
In addition to these, different approaches following the same themes are present in the literature. 
For instance, instance mapping using cosine similarity \cite{evaluating_wijekoon_2020} and label propagation on data clusters using radial basis function (RBF) kernel similarity \cite{semisupervised_presotto_2022} are explored. 
The application of an RBF kernel constraints the similarity measurements in the range between 0 and 1 by measuring distances in exponential terms. 
Instance weighting based on metrics like log-likelihood function \cite{importanceweighted_hachiya_2012} and meta classification loss \cite{swladapt_hu_2023} is also tested.
The meta classification loss is calculated by applying the standard loss functions on the transformed feature space for only the high-confidence pseudo labels generated by the classifier.  

There have also been some instance transfer works that follow themes different than the ones mentioned. 
For example, the metadata corresponding to physical, lifestyle, and sensor aspects of the source/target settings is used to create a similarity network, which is used in the boosting process \cite{enabling_lane_2011}.
Optimal source selection, in terms of the most compatible user, is another such theme. 
Meta features like predictability of activities using nearest neighbor classifier, Shannon entropy based diversity, frequency of activities, and pairwise dissimilarity between distributions of individuals are used in this selection process \cite{transfer_shen_2023}.

Similar to the case of instance transfer, several feature transfer approaches are used to address the challenges of personalization as well as feature heterogeneity. 
Such approaches include: (\textit{i}) conventional methods like TCA and SA \cite{transfer_li_2022, unsupervised_barbosa_2018} and (\textit{ii}) neural network induced feature space projections, guided by loss functions like domain discrimination/classification loss \cite{chakma2021activity,adversarial_suh_2022,domain_wilson_2022,domain_qin_2022,semisupervised_chakma_2022,strangan_faridee_2021}, MMD-based loss \cite{adversarial_suh_2022,domain_qin_2022,domain_hussein_2022,multisource_dong_2023,stratified_wang_2018}, triplet loss \cite{endtoend_ye_2021}, contrastive loss \cite{domain_wilson_2022}, and large margin loss \cite{semanticdiscriminative_lu_2022}.
Apart from these loss functions, the Jensen–Shannon divergence (JSD) loss is also used in intermediate feature space construction \cite{augtoact_faridee_2019} for personalization. 
It is a symmetric as well as bounded version of the KL divergence.  

Correlation alignment (CORAL) is a conventional feature transfer method used for personalization that minimizes the distance between the covariance matrices of the source and target feature vectors \cite{transfer_li_2022}.
Unlike CORAL, which works on source and target feature spaces, locally linear embedding (LLE) considers only the $k$-nearest neighbors around each data instance and tries to reconstruct each data point linearly using this neighborhood \cite{personalized_saeedi_2018}.  
In particular, LLE learns a weight matrix denoting reconstruction weights corresponding to each data instance pair. 
The corresponding optimization problem is solved using singular value decomposition (SVD) and eigenvector-based techniques. 
Latent Dirichlet allocation (LDA) is a hierarchical model that is commonly used for topic modeling, where documents are modeled by assuming a multinomial distribution over latent topics, which are further modeled using a similar distribution over words.
In the case of wearable sensor data, because of its multidimensional nature and continuous feature spaces, making the likelihood assumptions in LDA is not only infeasible but also prone to overfitting due to a large number of parameters. 
Thus, AdaBoost is used as proxy to calculate the predictive likelihood in LDA towards performing feature space transfer  \cite{learning_wen_2017,framework_wen_2017}.

In addition to these well-established methods, feature space transfer for personalization has also been achieved using a few customized approaches. 
One such approach performs supervised locality preserving projection (SLPP) to map source and target data on a common subspace such that the same class instances are mapped closer \cite{personalized_fu_2021}. 
This is achieved by creating an adjacency graph and solving an eigenvalue problem that incorporates smoothness and class-compactness terms.
Another approach applies a transformation on the source space that minimizes a scaled distance between projected source and target instances \cite{semisupervised_hashemian_2021}. 
The scaling factor in this case is the confidence of the classifier in predicting the activity label.
In addition to these transformation-based approaches, semantic feature space representation using activity attributes is also explored in the literature \cite{nuactiv_cheng_2013}. 
These attributes include key movement patterns like \textit{arm-up}, \textit{squat-stand}, etc.  

Under parameter transfer approaches, the \textit{pretrain-then-finetune} theme has been largely employed using neural network architectures like multi-layer perceptions (MLP) \cite{semisupervised_presotto_2022}, AE \cite{adversarial_suh_2022}, CNN \cite{cnn_chikhaoui_2018,personalized_rokni_2018,semisupervised_presotto_2022,transfer_an_2020,transnet_rokni_2020}, CNN-LSTM \cite{closedloop_saeedi_2017}, Bayesian CNN \cite{personalizing_akbari_2020}, and spatial transformers \cite{strangan_faridee_2021}.   
A variation to this theme by replacing the finetuning process with clustering-based label assignment is tested \cite{crosspeople_zhao_2011}. 
In the pretraining aspect, several training strategies like model agnostic meta-learning \cite{adapting_gong_2022, metasense_gong_2019}, soft parameter sharing \cite{domain_hussein_2022,metier_chen_2020}, domain adaptive batch normalization \cite{incremental_mazankiewicz_2020}, joint training \cite{metier_chen_2020}, and disentangled feature learning \cite{latent_qian_2021} are explored in the literature. 

Model agnostic meta learning presents an avenue, where different HAR tasks are constructed by sampling data across activities and subjects from the available datasets \cite{adapting_gong_2022, metasense_gong_2019}. 
Then, the target task is matched with one (or few) of the existing tasks using similarity measures based on mean distance measurements. 
Domain adaptive batch normalization is based on the fact that the batch normalization layer encodes domain-specific information. 
The idea of batch normalization is to keep the input distribution to each layer constant by replacing the output activations of previous layers with their standardized values. 
Thus, to perform transfer, instead of using exponentially weighted average to compute statistics, the entire test domain data is used to impose the same distribution across source and target domains \cite{incremental_mazankiewicz_2020}. 
Thus, the network can then produce representations in a subspace where both source and target distributions are similar. 

In the joint training framework, different tasks like HAR and subject recognition can simultaneously be optimized \cite{metier_chen_2020}. 
The subject recognition task enables the HAR model to extract subject-specific features that are essential for personalization. 
Disentanglement refers to learning features that correspond to underlying independent factors. 
One way of measuring disentanglement is via correlation, which is minimized \cite{latent_qian_2021}.
However, naively applying disentanglement methods to sequential models may intensify the KL vanishing problem. 
Thus, an evidence lower bound decomposition strategy for disentanglement is proposed in the literature \cite{towards_li_2022}.
It follows a multi-level disentanglement approach, covering both individual latent factors and group semantic segments. 

In addition to these, many conventional approaches like self-training \cite{improving_longstaff_2010}, co-training \cite{improving_longstaff_2010}, active learning \cite{improving_longstaff_2010,nuactiv_cheng_2013,online_sztyler_2017,personalizing_akbari_2020}, ensemble learning \cite{multisource_wilson_2020,personalized_reiss_2013,multisource_dong_2023}, and probabilistic Bayes-optimal classifiers \cite{active_diethe_2016} are also used for personalization. 
In the active learning frameworks, the selection of data instances for training is performed based on least confidence \cite{nuactiv_cheng_2013}, margins in predictions \cite{nuactiv_cheng_2013}, maximum entropy \cite{nuactiv_cheng_2013}, and user feedback \cite{online_sztyler_2017}.
Among the ensemble-based methods, techniques are explored that target training of the ensemble, as well as inference based on it. 
In addition to setting-specific training of the individual classifiers, their loss functions can also be complemented with KL divergence between the label distributions of source and target domains to perform domain adaptation \cite{multisource_wilson_2020}. 
Another strategy to perform additional training of the ensemble is to update only the weights assigned to each individual component based on unseen data \cite{personalized_reiss_2013}. 
For inference in ensembles, techniques like majority voting \cite{personalized_reiss_2013} and PCR5 \cite{multisource_dong_2023} are commonplace. 
A novel inference technique, called calibration \cite{toward_hong_2016}, is designed for learning a personal model, where the goal is to select the best-performing individual classifiers per activity or subject to match the target conditions. 
Under probabilistic Bayes-optimal classifiers, a hierarchical multi-class Bayes point machine is used for performing transfer from a community to an individual subject \cite{active_diethe_2016}. 
A Bayes point machine learns $k$ linear discriminant functions for a $k$-class classification problem, where the goal is to minimize average error marginalized over the entire sampling of data instances. 
Under the transfer learning setup, the prior values of this model, originally over the community weights, are replaced by Gaussian and Gamma posteriors learned from individual-specific data.

\begin{table}
\caption{Exploration of transfer learning challenges of multi-source training, dynamic environments, task difference, distributed/federated setups, and performance evaluations in wearable HAR literature with corresponding source-target dataset pairs:
We categorize the transfer learning works in terms of the challenges they address and the source-target dataset pairs on which their experiments are performed. 
In this table, the "Others" dataset category denotes custom datasets collected specific to some works as well as the lesser-used publicly available datasets like MotionSense, HASC, MyoGym, USC-HAD, Wetlab, Skoda, UniMiB, uWave, COSAR, KU-HAR, SPAD, Bilkent, Daphnet, WHARF, RealDisp, and Capture 24.
}
\label{tab:wearable_datasets_challenges_2}
\begin{adjustbox}{max width=\textwidth} 
  \begin{tabular}{M{2.5cm}M{5.5cm}M{5.5cm}M{5.5cm}M{5.5cm}}
    \toprule
     Datasets & Multi-Source Transfer & Dynamic Environments & Difference in Tasks & Performance Measurement \\
    \midrule
    HHAR & \cite{multisource_wilson_2020, combining_presotto_2023,domainrobust_zhao_2022} & \cite{adarc_roggen_2013} & \cite{haresamudram2022assessing,dhekane2023much} & \cite{haresamudram2022assessing,dhekane2023much} \\
    PAMAP2 & \cite{multisource_dong_2023, chakma2021activity,combining_presotto_2023,domainrobust_zhao_2022,recurrenthar_kumar_2022,semisupervised_chakma_2022,tasked_suh_2022} & \cite{novel_hu_2019,deep_yan_2022} & \cite{haresamudram2022assessing,dhekane2023much} & \cite{investigating_hasthanasombat_2022,haresamudram2022assessing,dhekane2023much} \\
    OPPORTUNITY & \cite{chakma2021activity,domainrobust_zhao_2022,semisupervised_chakma_2022,tasked_suh_2022} & \cite{adaptive_wen_2016,adaptive_abdallah_2015,streamar_abdallah_2012} & \cite{novel_fan_2022,crossdomain_khan_2022,transfer_du_2019,untran_khan_2018,unlabeled_bhattacharya_2014} & - \\
    DSADS & \cite{multisource_dong_2023, chakma2021activity,semisupervised_chakma_2022} & \cite{novel_hu_2019} & \cite{smoke_nguyen_2015} & - \\
    Mobiact & \cite{combining_presotto_2023} & - & \cite{haresamudram2022assessing} &  \cite{haresamudram2022assessing} \\
    RealWorld & \cite{combining_presotto_2023} & \cite{novel_hu_2019} & \cite{dhekane2023much} & \cite{dhekane2023much} \\
    UCI & \cite{multisource_wilson_2020, combining_presotto_2023} & \cite{crossperson_deng_2014} & \cite{transact_khan_2017} & - \\
    MHealth & \cite{tasked_suh_2022} & - & \cite{deep_yan_2022,recognizing_nguyen_2015,haresamudram2022assessing}  & \cite{investigating_hasthanasombat_2022,haresamudram2022assessing} \\
    WISDM & \cite{multisource_wilson_2020, domainrobust_zhao_2022,recurrenthar_kumar_2022} & \cite{adaptive_abdallah_2015} & \cite{crossdomain_khan_2022,deeptranshar_kumar_2023,untran_khan_2018} & - \\
    DAS & - & - & \cite{crossdomain_khan_2022,recognizing_nguyen_2015,untran_khan_2018} & - \\
    Custom & 
    \cite{tasked_suh_2022, recurrenthar_kumar_2022, multisource_wilson_2020, domainrobust_zhao_2022, enabling_lane_2011,combining_presotto_2023} & \cite{autonomous_rokni_2018, adarc_roggen_2013,streamar_abdallah_2012,adaptive_abdallah_2015} & \cite{deeptranshar_kumar_2023, haresamudram2022assessing, deep_yan_2022,nuactiv_cheng_2013,unlabeled_bhattacharya_2014,transfer_du_2019} 
    & \cite{haresamudram2022assessing,investigating_hasthanasombat_2022} \\
\bottomrule
\end{tabular}
\end{adjustbox}
\end{table}

\subsection{Multi-Source Transfer Learning}
One of the biggest challenges in the domain of wearable HAR is the unavailability of large-scale datasets, where generalized models can be learned and deployed across diverse settings. 
One way to address this is to train the HAR models on a number of small-scale datasets. 
However, based on the previous discussions, the factor of heterogeneity induced by sensor modalities, sensing positions, device aspects, and human subjects makes it difficult to leverage these datasets in a straightforward manner. 
Thus, the research on using multiple source datasets involves scalable feature transfer techniques that can handle multiple (number and type of) heterogeneities at the same time, while avoiding negative transfer.  
To address this challenge, several works have used source-target pairs based on the datasets shown in Table \ref{tab:wearable_datasets_challenges_2}. 

\subsubsection{Approaches}
Since most of the conventional feature transfer techniques (e.g., TCA, SA, and CORAL) are designed for a single-source transfer learning setup, the use of neural network based training frameworks is preferred for multi-source feature transfer. 
This includes the exploration of various architectures (e.g., ResNet, DeepConvLSTM, HART, and MobileHART) \cite{combining_presotto_2023}, loss functions (e.g., domain discrimination loss \cite{chakma2021activity,multisource_wilson_2020,semisupervised_chakma_2022,tasked_suh_2022}, MMD-based loss \cite{multisource_dong_2023,tasked_suh_2022}, and self-distillation loss \cite{tasked_suh_2022}), and training strategies \cite{combining_presotto_2023,domainrobust_zhao_2022,recurrenthar_kumar_2022}.  
While these frameworks aim to arrive at a common feature space, they achieve it by sharing parameters (i.e., having common feature extractors), which can also be seen as a parameter transfer method. 
The loss functions like MMD and domain discrimination loss are calculated between each pair of source and target domains and used to formulate a global loss function that is jointly optimized. 
The self-distillation loss is applied in a teacher-student framework.
In that, a global model is pre-trained that acts as the teacher, and the KL divergence between the predictions of this teacher and individual student model is used as a loss function \cite{tasked_suh_2022}.  

A training strategy similar to cross-validation, called leave-one-dataset-out, is explored, where all the datasets except one are used for pre-training followed by finetuning on the remaining one \cite{combining_presotto_2023}. 
For handling differences across domains in terms of their tasks (activity label sets), the pre-training can be performed by classifying the target activities into \textit{basic} and \textit{complex} types \cite{domainrobust_zhao_2022}. 
The data corresponding to the basic activities, which are common across domains, can then be used to perform pre-training.   
In standard pre-training setups, it is often assumed that all the layers of the neural network architecture are updated. 
However, in a multi-source setup, this policy can incur significant costs, which can be reduced by performing standard pre-training on one source dataset followed by updating only the classification layers for the remaining ones \cite{recurrenthar_kumar_2022}. 
In neural network based multi-source frameworks, the features corresponding to individual source domains can be concatenated to deliver a single inference by the HAR model. 
However, it is also possible to build an ensemble framework by considering individual source classifiers as building blocks. 
In the literature, one such technique is explored, where a weighted combination of individual source classifiers based on the perplexity scores between features is used to perform inference \cite{chakma2021activity}.

Apart from feature and parameter transfer techniques, an instance transfer technique based on similarity-sensitive boosting is also used in a multi-source setup \cite{enabling_lane_2011}. 
The similarity between instances is calculated using similarity networks based on physical, lifestyle, and sensor-related metadata.

\subsection{Difference in Tasks}
The challenge of task difference across domains refers to a change in the activity label set, i.e., $Y^s \neq Y^t$, or their conditional distribution, i.e., $P(Y^s|X^s) \neq P(Y^t|X^t)$. 
The origin of this problem lies in the data collection protocols, where the data instances are annotated with only a specific set of activities, for example, \textit{biking}, \textit{sitting}, \textit{stairs up}, \textit{stairs down}, \textit{standing}, and \textit{walking}. 
This process results in two main issues. 
First, the human subjects perform many additional movements, in addition to the activity with which the data instances are annotated. 
In some cases, this results in the data instances for the same activities being very different \cite{plotz2023if}.
Second, some activities like \textit{walking} and \textit{nordic walking}, while being different, are very similar to each other.
Thus, to perform a transfer where activity sets differ, it is important to learn data representations that are invariant to the activity labels. 
Similar to the case of task difference in smart home settings, there exists an implicit assumption that the activities across domains are similar in terms of semantics or the underlying motion patterns.
To incorporate task differences across domains, several works have used source-target pairs based on the datasets shown in Table \ref{tab:wearable_datasets_challenges_2}. 

\subsubsection{Approaches}
The main requirement for performing a transfer across different tasks is to have learned more generalized activity-invariant data representations.  
One way of achieving this goal is to have a feature representation that is suitable for such transfer. 
Semantic feature representation is one such way in which data can be represented \cite{nuactiv_cheng_2013,towards_cheng_2013}. 
Semantic attributes can be seen as activity descriptors, e.g., \textit{arm up}, \textit{sitting on table}, etc. 
Since the underlying assumption in this transfer ensures the activities across domains are \textit{similar}, common semantic attributes are more likely to exist, which can enable the transfer learning process. 
While semantic space representation relies on the manual designing of relevant attributes, neural networks are capable of extracting transferable features without any hard coding.
AE \cite{crossdomain_khan_2022,untran_khan_2018}, VAE \cite{novel_fan_2022}, GNN \cite{deep_yan_2022}, and models based on convolutional and recurrent layers like DeepTransHAR \cite{deeptranshar_kumar_2023} and DeepConvLSTM \cite{transfer_du_2019} are used in the literature to extract such features. 

In addition to these broader approaches, specialized methods are also explored in the literature to tackle the problem of task difference.  
For instance, the multi-class task difference problem is simplified into multiple one-vs-all binary classification problems \cite{smoke_nguyen_2015}. 
The inference is then performed by combining the predictions of individual classifiers based on the highest confidence measure.  
This approach helps in detecting unseen classes based on low confidence values among all binary classifiers while avoiding false positives.
Unseen classes during transfer can also be recognized using anomaly detection followed by clustering \cite{transact_khan_2017}. 
Anomaly detection helps in identifying the \textit{negative} instances that may represent unseen activities. 
Performing clustering using such negative instances followed by label propagation using a majority voting based approach results in an instance transfer approach designed to identify similar as well as unseen activities across domains.

The idea of semantic space representation is extended in the literature \cite{recognizing_nguyen_2015}, where classifiers are trained on the existing as well as the semantic feature space. 
In case the target data instance belongs to an unseen class, the attribute-based classifiers are used. 
A slight variation in semantic space representation is explored in the activity space via partonomy-based transfer \cite{remember_blanke_2010}, where activities are broken into components that can be combined to form unseen activities. 
To do so, the data instances are segmented at low-movement points followed by the extraction of statistical features and training of a conditional random field (CRF) model to detect composite activities. 
Another unsupervised representation learning approach useful in this scenario is sparse coding \cite{unlabeled_bhattacharya_2014} that tries to learn an over-complete set of feature vectors.
This is performed by formulating an optimization problem based on reducing the reconstruction error and increasing the sparsity in the learned vectors in the codebook learning phase.
Later, a classifier is trained by identifying clusters within the learned codebook and selecting the most informative basis vectors from the cluster to learn the classifier.   
Self-supervised learning has also been explored to learn task-independent features \cite{haresamudram2022assessing}. 
In the pretraining step, it constructs a \textit{pretext task} using the unlabeled data itself to learn features that are later reused in the finetuning step. 

The use of other data modalities for recognizing the unseen activities has been explored in the literature. 
In particular, the features extracted from IMU data can be used for generating synthetic 2D skeletal maps in an adversarial framework, where the inference can be performed on the skeletal data using pre-trained models \cite{novel_fan_2022}.
This approach leverages the availability of data and pre-trained models from another domain to complement sensor-based transfer learning for HAR.

\subsection{Dynamic Environments}
The standard transfer learning setup assumes a change in the feature space or task across domains only once.
Thus, most of the transfer learning frameworks are trained only once to address the feature space heterogeneity, user variation, or difference in tasks. 
However, in real-world scenarios, a target environment is likely to undergo a change in factors like sensor positions (smartphone from left to right pocket), target users, task differences, etc. 
To perform HAR in such dynamic environments, transfer learning needs to be performed iteratively using methods that are capable of quickly adapting to the new target setting. 
A typical framework for performing such a transfer involves training a base model and updating it periodically with encountering a change in domain. 
In the literature, several works have addressed this challenge using source-target pairs based on the datasets shown in Table \ref{tab:wearable_datasets_challenges_2}.
Below, we present a subset of approaches addressing this challenge. 
A more comprehensive survey of transfer learning methods for dynamic environments is presented by Abdallah et al.\ \cite{activity_abdallah_2018}.  

\subsubsection{Approaches}
Different kinds of heterogeneities are introduced in dynamic environments, which are usually unknown at the initial model-building stages. 
Thus, it is essential for the model to work in a representation space that can easily accommodate different sensor heterogeneities, distribution changes, and other variations. 
Semantic space \cite{adaptive_wen_2016} and clustering-based representation \cite{adaptive_abdallah_2015,autonomous_rokni_2018,streamar_abdallah_2012} are popular choices for constructing such domain-invariant representation space.   
Semantic space representation uses commonly present attributes related to sensor type, data type, sensor location, attached object, activity patterns, etc., as features of the data instances.
Clustering, on the other hand, can be separately performed on different feature spaces. 
Then, the resultant clusters can directly be mapped with each other to bypass the construction of an intermediate feature space. 
Similar to clusters, subsequences among data instances can also be mapped based on similarity measures, provided they do not belong to very different feature spaces \cite{transfer_saeedi_2016}. 

After addressing the challenge of data representation, performing inference \& updating the base model is the next step.
In the frameworks where clustering-based data representation is present, the inference can be performed using active learning \cite{adaptive_abdallah_2015,streamar_abdallah_2012}, weighted graph matching \cite{autonomous_rokni_2018}, and cluster updation techniques \cite{streamar_abdallah_2012}.   
An inference technique invariant of the underlying data representation scheme is to update the classifier in a self-training framework, where a new training dataset is iteratively created based on the confident predictions of the base model on the unseen data \cite{crossperson_deng_2014,adarc_roggen_2013}.
In addition to dynamically creating training datasets, another approach is to modify the classifier itself with a change observed in the observed setting. 
This can be performed by using feature incremental random forests \cite{novel_hu_2019}, where a novel mutual information based diversity generation strategy is employed to select individual trees of the random forests that become ineffective with the emergence of new sensors. 
To incorporate new data, steps like the creation of new trees and the update of older trees by performing node splitting on leaves can be performed. 
The adaptation process can also be viewed as a learning-to-rank problem \cite{adaptive_wen_2016}, where the goal is to rank activity classes for each data instance. 
This can be performed by computing a score against each activity class for each data instance using semantic attributes and maximizing the area under curve (AUC) using stochastic gradient descent.

\subsection{Distributed \& Federated Learning}
In wearable devices like smartwatches and smartphones, movement-based (IMU) and physiological (heart rate, blood sugar, etc.) sensors are deployed that are capable of constantly collecting user-specific data. 
Hypothetically, the collection of this data from multiple users can likely result in a large-scale dataset, on which, generalizable HAR models can be trained. 
However, these data modalities are often an indicator of the user's health, and hence, sharing this data would violate privacy as well as security.   
Recently, there have been regulations put in place to protect data privacy and security \cite{voigt2017eu}.  
Hence, the data is now available in the form of isolated islands \cite{fedhealth_chen_2020}, which makes it impossible to perform standard training of HAR models in a centralized manner. 
Along with the data availability, there are additional challenges of resource constraints and personalization. 
Since the model training has to be shifted onto small-scale devices, handling computational resources becomes imperative. 
Also, the HAR system on each small-scale device needs to be personalized to the end user. 
All these challenges combined give rise to the research area of federated/distributed transfer learning.   
To perform simulated experiments, datasets like HHAR, Mobiact, and UCI Smartphone are used in the literature. 

\subsubsection{Approaches}
A typical federated/distributed learning setup consists of a server and a client side. 
The goal of the client side is to perform on-device setting-specific training, whereas, the server side is responsible for \textit{combining the client models} to create a global model, which benefits the client models as they have less training data.   
On the server side, the availability of computational resources is greater than on the client side, however, the training data is not available as well. 
Thus, there has to be communication between both sides in terms of sharing the model weights, classification scores, or other training-specific characteristics that do not violate any privacy or security policies. 
Under these constraints, federated learning frameworks are set up. 

Since the server side can afford the training of neural networks, the \textit{pretrain-then-finetune} theme is explored in the federated learning setup \cite{fedhealth_chen_2020,semisupervised_presotto_2022}. 
On the server side, the first step is to pretrain the initial HAR model using publicly available datasets. 
These models are shared with the clients for them to bootstrap their personal HAR system. 
These models are then re-trained on the client side using small amounts of personal data. 
These updated models are shared back with the server, where they are aggregated to update the global model.
The updated global model gets shared with the clients again and the entire process is repeated. 

In addition to sharing the model weights, the class-wise prediction scores can also be shared between the server and the clients \cite{federated_gudur_2020}. 
The benefit of sharing class-wise probabilities is that it becomes straightforward to use a knowledge distillation scheme on the client side, where having small-size models is a requirement.  
Under a distributed learning setup, a workaround for sharing data has been explored \cite{semisupervised_hashemian_2021}, where a privacy-protecting dataset is created on the server side.
Rather than directly sharing data instances, which would violate privacy, all clients agree to a specific matrix that gets applied to the user-selected data instances to generate an encoded data matrix, which is shared with the server.

\subsection{Performance Measurement}
The standard evaluation setup in machine learning consists of $k$-fold cross-validation. 
The idea behind using it is to avoid overfitting and potential bias in the training data. 
Additionally, performance indices like weighted F1 score or AUC are preferred over standard classification accuracy 
as they are more suitable under class imbalance settings \cite{mullick2020appropriateness}.  
However, all these remedies are ineffective at measuring the performance of a HAR framework when the train and test conditions differ drastically.
In particular, measuring the overall performance of a HAR framework based only on its performance in the source setting does not provide a complete picture of the ability of the framework to generalize. 
Thus, performance measurement under transfer learning settings is crucial for assessment and comparison purposes. 
The source-target dataset pairs used in performance measurement works are shown in Table \ref{tab:wearable_datasets_challenges_2}.

\subsubsection{Approaches}
The works targeting performance evaluation under transfer learning settings are very few. 
In a multi-domain setup, a \textit{leave-one-dataset-out} based performance measure is used for analyzing the domain-agnostic performance of a classifier \cite{investigating_hasthanasombat_2022}. 
Similar to $k$-fold cross-validation, it performs $k$ trials of pre-training, where $k-1$ datasets are used as sources and inference if performed on the remaining dataset. 
A more comprehensive assessment targeted at characterizing model performance based on the robustness to differing source and target conditions, the influence of dataset characteristics, and the feature space characteristics is performed in the literature \cite{haresamudram2022assessing}. 
This large-scale assessment work focuses on seven self-supervised approaches, namely, multi-task self-supervision, masked reconstruction, AE, CPC, SimCLR, SimSiam, and BYOL, and analyzes them from the perspectives of the aforementioned factors on ten datasets.  
Another work focused on analyzing self-supervised frameworks from the perspective of the pre-training data efficiency has proposed an index called \textit{critical mass}. 
It denotes the minimum amount of unlabeled data used for pre-training, which results in a recognition F1 score within 5\% of using the entire dataset.

%% file: discussion.tex
\subsection{Review of the State-of-the-art}
\label{subsec:review_sota}


\begin{table}
\caption{Review of the state-of-the-art works in the domain of transfer learning in smart home settings for HAR. 
We shortlist those works that claim their approach to be the best-performing one by comparing it against other relevant methods in the literature. 
We also filter out the \textit{then-SOTA} methods, i.e., the ones that were the best-performing at the time they were proposed but are no longer SOTA.
}
\label{tab:smart_home_sota}
\begin{adjustbox}{max width=\textwidth} 
  \begin{tabular}{M{2.5cm}M{1cm}M{2.5cm}M{2.5cm}M{7cm}M{5cm}M{2cm}M{2cm}}
    \toprule
     Reference & Year & Datasets & Tasks & Method & Compared Against & Performance Metric & Performance (\textit{min - max}) \\
    \midrule
    Sukhija \& Krishnan \cite{sukhija2019supervised} & 2019 & CASAS HH & Different Tasks, Heterogeneity & Random forests to identify pivotal features from both domains to learn a transformation matrix & SHFR, HeMap, FA, HFA, SHFR-RF & mean error, standard deviation & $26.49 \pm 2.81$ - $18.17 \pm 1.68$ \\
    Feuz \& Cook \cite{feuz2015transfer} & 2015 & Custom & Heterogeneity, Multi-Source Transfer & Meta feature construction for heterogeneity and stacking in ensembles for multiple sources & Ablations & accuracy, recall & accuracy $\sim 0.5 - 0.62$, recall $\sim 0.2 - 0.3$ \\
    Sanabria et al. \cite{sanabria2021unsupervised} & 2021 & University of Amsterdam & Heterogeneity & Shift-GAN, a Bi-GAN variant with covariate shift correction using KMM, for feature transfer & GFK, TCA, FLDA, JDA, IW, CCA, ADDA, DAN, DANN, ADADM & macro F1 score & 0.68 - 0.76 \\
    Ramamurthy et al. \cite{ramamurthy2021star} & 2021 & Custom, OPPORTUNITY & Different Tasks & Pretrain-then-finetune: self-taught feature extraction using deep belief networks & w/o self-taught learning in nearest neighbor, SVM, RF, MLP, CNN & accuracy, F1 score & F1 score: OPPORTUNITY: 0.86 \\
\bottomrule
\end{tabular}
\end{adjustbox}
\end{table}

\begin{table}
\caption{Review of the state-of-the-art works in the domain of transfer learning in wearables for HAR.
We observe a significant number of works claiming their approach to be the SOTA. 
Hence, we first select only those approaches that were proposed in the last three years, i.e., from 2021 to 2023. 
Similar to the case of smart homes, we filter out the \textit{then-SOTA} methods.  }
\label{tab:wearable_sota}
\begin{adjustbox}{max width=\textwidth} 
  \begin{tabular}{M{2.5cm}M{1cm}M{2.7cm}M{2.5cm}M{7cm}M{5cm}M{2.5cm}M{2.7cm}}
    \toprule
     Reference & Year & Datasets & Tasks & Method & Compared Against & Performance Metric & Notable Performance \\
    \midrule
    Gong et al. \cite{adapting_gong_2022} & 2022 & Custom & Personalization & Model agnostic meta learning: different HAR tasks constructed by sampling data across activities and subjects & Transfer with CNN, Prototype Networks, MAML & average accuracy, AUC & accuracy $\sim 0.8 - 0.91$, AUC 0.98  \\
    Suh et al. \cite{adversarial_suh_2022} & 2022 & OPPORTUNITY, PAMAP2, MHEALTH, MoCapci & Personalization & Training of AE based on MMD and adversarial subject-classification loss & Multi-channel CNN, DeepConvLSTM, Transformers, METIER & accuracy, weighted \& macro F1 score & macro F1 score: MHEALTH $96.47 \pm 4.04$, PAMAP2 $77.84 \pm 11.69$   \\
    Presotto et al. \cite{combining_presotto_2023} & 2023 & HHAR, MobiAct, PAMAP2, RealWorld, MotionSense, UCI & \textit{d}-Heterogeneity & multi-dataset training in leave-one-dataset-out manner & Inception-ResNet, DeepConvLSTM, HART, MobileHART architectures & average F1 score & Inception-ResNet: PAMAP2: 0.76, MotionSense: 0.97 \\
    Sanabria et al. \cite{contrasgan_sanabria_2021} & 2021 & PAMAP2, DSADS, WISDOM & \textit{p}-Heterogeneity, Personalization & Bi-GAN for feature alignment and contrastive learning & TCA, GFK,  DAN, DANN, JAN, DeepCORAL, ADADM & micro \& macro F1 score & macro F1 score: PAMAP2: 0.76, DSADS: 0.84\\
    Lu et al. \cite{crossdomain_lu_2021} & 2021 & DSADS, UCI-HAR, USC-HAD, PAMAP2 & \textit{p}-Heterogeneity & mapping between substructures of data using optimal transport method & nearest neighbor, TCA, SA, CORAL, STL, OT & accuracy & PAMAP: 0.74, DSADS: 0.81 \\
    Khan \& Roy \cite{crossdomain_khan_2022} & 2022 & OPPORTUNITY, WISDOM, DAS & Different Tasks & AE trained using cross-entropy and MMD loss & UnTran, JDA, TCA & accuracy & WISDM: 0.89, DAS: 0.89 \\
    Zhu et al. \cite{dmstl_zhu_2023} & 2023 & PAMAP2, DSADS, SHL & \textit{p}-Heterogeneity & source domain selection: MMD-based distances, feature transfer: multi-head attention, local MMD loss, domain confusion \& classification loss & TCA, JAN, TNNAR, DANN, SymNet, UDA-FM & accuracy, macro-F1 score & macro-F1 score: PAMAP2: 0.42, DSADS: 0.73 \\
    Alajaji et al. \cite{domain_alajaji_2023} & 2023 & Custom & \textit{p}-Heterogeneity & CNN \& soft attention module trained with domain alignment, triplet, and classification loss & CORAL, DNN, HDCNN & accuracy, micro-F1 score & micro-F1 score: 0.54 \\ 
    Hussein \& Hajj \cite{domain_hussein_2022} & 2022 & PAR, HHAR & \textit{d}-Heterogeneity, Personalization & soft parameter sharing \& squared MMD loss & Hard \& soft parameter sharing, linear \& non-linear relations, DC \& MMD loss & average F1 score & HHAR: 0.84  \\
    Soleimani et al. \cite{generic_soleimani_2022} & 2022 & PAMAP2, OPPORTUNITY, LISSI & Personalization & Synthetic data generation: CNN-based GAN with micro-mini batch training strategy & STL, GFK, SA-GAN, DANN, VADA, IIMT & weighted-F1 score & PAMAP2: 0.76, OPPORTUNITY: 0.54  \\
    Varshney et al. \cite{human_varshney_2021} & 2021 & mHealth, PAMAP2 & \textit{p}-Heterogeneity & pretrain-then-finetune: CNN backbone & PerceptionNet, DeepConvLSTM, Stochastic Generative Network & accuracy & PAMAP2: 0.94, mHealth: 0.98 \\
    Qian et al. \cite{latent_qian_2021} & 2021 & UCI-HAR, UniMiB, OPPORTUNITY & Personalization & disentangled feature learning using VAE with independent excitation mechanism & DDNN, DeepConvLSTM, CoDATS, VAE, DIVA & F1 score & UniMiB: 0.88, OPPORTUNITY: 0.81  \\
    Zhao et al. \cite{local_zhao_2021} & 2021 & DSADS, PAMAP2 & \textit{p}-Heterogeneity, Personalization & cluster-to-cluster transfer using MMD between clusters & Nearest Neighbor, SVM, RF, PCA, TCA, SA, STL & accuracy, F1 score & PAMAP F1 score: \textit{p}-Heterogeneity: 0.72, Personalization: 0.4 \\
    Alinia et al. \cite{modelagnostic_alinia_2023} & 2023 & PAMAP2, DAS, Smartsock & \textit{p,m}-Heterogeneity & combinatorial optimization: dependency graph with min-cost mapping & ConvLSTM, DirectMap & F1 score, NMI, Purity & PAMAP2 F1 score: \textit{m}-heterogeneity: 0.59, \textit{p}-heterogeneity: 0.7 \\
    Fallahzadeh et al. \cite{personalized_fallahzadeh_2023} & 2023 & DSADS, PAMAP2, Smartsock & Personalization & Similarity network based clustering followed by bipartite mapping & CvC, CvR, RF, Bagging, ActiLabel & accuracy & PAMAP2: $\sim 0.74$, DSADS: $\sim 0.72$ \\
    Lu et al. \cite{semanticdiscriminative_lu_2022} & 2022 & SHAR, DSADS, UCIHAR, PAMAP2, USCHAD & \textit{p}-Heterogeneity, Personalization & Semantically aware augmentations and large margin loss for feature transfer & DANN, CORAL, ANDMask, GroupDRO, RSC, GILE & accuracy, standard error & PAMAP2: 0.81, DSADS: 0.91 \\ 
    Faridee et al. \cite{strangan_faridee_2021} & 2021 & PAMAP2, OPPORTUNITY, HHAR & \textit{p}-Heterogeneity, Personalization & spatial transformer with domain discrimination loss & STL, MADA, DGDA, SA-GAN & average F1 score PAMAP2 & \textit{p}-Heterogeneity: 0.61, Personalization: 0.7   \\
    Hu et al. \cite{swladapt_hu_2023} & 2023 & SBHAR, OPPORTUNITY, RealWorld & Personalization & Instance weighting based on classification loss, domain discrimination loss, and meta-classification loss & HDCNN, MMD, dan, AdvSKM, MCD, XHAR, DANN, DUA, ETN, TCL, UAN, SS-UniDA, PADA & accuracy, macro F1 score & macro F1 score: OPPORTUNITY: 0.58, RealWorld: 0.74 \\
    Suh et al. \cite{tasked_suh_2022} & 2022 & OPPORTUNITY, PAMAP2, MHealth, RealDisp & Multi-Source & pretrain-then-finetune: backbone with spatial attention block trained using MMD, domain discrimination, and self-knowledge distillation loss & MC-CNN, DeepConvLSTM, Self-Attention, METIER, Adversarial CNN & accuracy, weighted \& macro F1 score & macro F1 score: PAMAP2: 0.75, OPPORTUNITY: 0.77  \\
    Pavliuk et al. \cite{transfer_pavliuk_2023} & 2023 & KU-HAR, UCI-HAPT & \textit{d}-Heterogeneity & pretrain-then-finetune: pre-trained CNN on scalogram generated using CWT & handcrafted, FFT, CNN-GRU, Wavelet Packet Transform based features & accuracy, F1 score & F1 score KU-HAR: 0.97 \\
    Haresamudram et al. \cite{haresamudram2023investigating} & 2023 & HHAR, Myogym, Mobiact, Motionsense, MHEALTH, PAMAP2 & \textit{p}-Heterogeneity, Different Tasks & Self-supervised learning: Enhanced CPC with modifications in encoder architecture, autoregressive network, and future prediction task & CNN, GRU, DeepConvLSTM, Multi-task SSL, AE, SimCLR, CPC & macro F1 score & HHAR: 0.59, MotionSense: 0.89, MHEALTH: 0.53 \\
    Leng et al. \cite{leng2023generating} & 2023 & PAMAP2, RealWorld, USC-HAD & \textit{m}-Heterogeneity & Synthetic IMU data generation using LLMs and CLIP based pre-trained models & Classification setup with real IMU data & macro F1 score & PAMAP2: 0.69, RealWorld: 0.77, USC-HAD: 0.48 \\
    
\bottomrule
\end{tabular}
\end{adjustbox}
\end{table}

In the last two sections, we surveyed $205$ works that perform transfer learning the domains of smart homes and wearables for HAR. 
These works range in the span of the last 16 years, where enhancements, in both the approaches and the performance, can be witnessed periodically.
With the advent of deep learning, the conventional transfer learning algorithms are somewhat replaced by neural network based solutions, where advancements are being made primarily in the neural network architectures, design of loss functions, and training frameworks.  
Thus, it is essential to keep track of the best-performing methods such that the community can benefit by learning the recent trends. 

For this purpose, we provide a review of the state-of-the-art (SOTA) transfer learning methods for HAR in smart home and wearable sensing scenarios in Table \ref{tab:smart_home_sota} and Table \ref{tab:wearable_sota}, respectively. 
In particular, we provide the following information about the SOTA approaches: (\textit{i}) references, (\textit{ii}) year, (\textit{iii}) datasets used in the experiments, (\textit{iv}) challenges tackled (from Section \ref{sec:smart_home} and Section \ref{sec:wearables}), (\textit{v}) Method(s) used to address these challenges, (\textit{vi}) methods with which the SOTA approaches are compared, (\textit{vii}) performance metrics used in their results, and (\textit{viii}) performance values. 
To shortlist the SOTA works, we first search for articles that claim to have the best-performing method(s) present. 
Then, we filter out the works that were proposed in the distant past (more than 3 years in the case of wearables literature).
We do not perform this filtering in the case of smart homes because of the lesser number of SOTA contributions. 
Finally, we remove the redundant entries in our collection, where clear extensions of these works have been proposed in the literature. 

Based on Table \ref{tab:smart_home_sota} and Table \ref{tab:wearable_sota}, we draw the following conclusions: (\textit{i}) The domain of transfer learning for HAR in smart homes has significantly less number of works that compare their proposed method with other relevant SOTA methods, (\textit{ii}) Around 73\% (19 out of 26) of the SOTA methods rely on deep learning based frameworks, (\textit{iii}) Commonly found themes in the SOTA deep learning based frameworks are \textit{pretrain-then-finetune}, architectures based on AE, CNN, GRU/LSTSM, GAN, and attention modules, loss functions based on MMD and domain discrimination loss, training frameworks like self-supervised learning, disentangled learning, and adversarial learning, (\textit{iv}) Commonly found themes in the SOTA conventional approaches are data representation using meta features, clustering approaches, and subsequences, label propagation using optimal transport and graph matching, (\textit{v}) Unless explicitly performed, a direct comparison of these SOTA methods is not possible owing to a variation in the experimental settings in terms of the datasets, challenges addressed, and performance metrics. 

\subsection{Roadmap for Future Work}
In Section \ref{sec:smart_home} and Section \ref{sec:wearables}, we categorized and presented the transfer learning approaches based on the challenges they address in the application domains of smart homes and wearables, respectively. 
Based on that discussion, one can observe that many challenges like feature space heterogeneity, personalization, multi-source transfer, etc., are common in both application domains, despite their differences in the problem settings.
Therefore, several transfer learning approaches can be applied in both these application domains to address similar kinds of challenges. 
For example, adversarial training using GANs is successfully used in both smart homes as well as wearable HAR to address the challenges of personalization and feature space heterogeneity (See Table \ref{tab:smart_home_sota} and Table \ref{tab:wearable_sota}). 
However, from Table \ref{tab:solution_space}, Table \ref{tab:smart_home_sota} and Table \ref{tab:smart_home_datasets_challenges}, one can observe that a significantly less amount of work has been done for solving the respective transfer learning challenges in smart home settings.
Moreover, there have been recent advances in the domains of computer vision and natural language (described later), from which, inspiration can be drawn for solving similar challenges in both the wearable and the smart home domains.

To systematically point out gaps in the literature and propose a roadmap to address them, we present the \textit{solution space} explored in the domains of smart homes and wearable HAR for transfer learning. 
We refer to the solution space as the different types of approaches explored in the relevant literature. 
A comprehensive picture of the solution space for smart homes and wearable application domains in terms of instance, feature, and parameter transfer techniques is presented in Table \ref{tab:solution_space}. 
Referring to these solution spaces and the state-of-the-art works from both domains, we identify the following research gaps and present roadmaps for future works to address them.

\begin{table}
\caption{Exploration of the solution space in the domains of smart homes and wearable HAR in terms of instance, feature space, and parameter transfer techniques. 
Most of the transfer learning methods, explicitly or implicitly, use domain-specific knowledge bases in terms of heuristics, context information, and metadata. Also, knowledge-based transfer methods, on their own, are not capable of performing transfer learning in the case of HAR. 
Instead, they are used as a supporting component in the instance, feature space, or parameter transfer frameworks.
Thus, we focus on the solution spaces in terms of these three types of transfer learning methods. 
We also note that many approaches in the Feature and Parameter transfer can be classified under both these categories, hence, we avoid this repetition by logging them under only the most relevant category. 
}
\label{tab:solution_space}
\begin{adjustbox}{max width=\textwidth} 
  \begin{tabular}{M{3.5cm}M{8cm}M{12cm}}
    \toprule
     Transfer Learning Type & Smart Homes & Wearables  \\
    \midrule
    Instance Transfer & 
    \textbf{Instance Mapping \& Label Propagation:} \newline Graph Matching \cite{chiang2017feature, chen2017activity, lu2014instantiation, chiang2012knowledge, dridi2022transfer}, EM framework \cite{rashidi2011activity, rashidi2010multi}, Clustering \cite{polo2020domain,alam2017unseen}, Activity templates \cite{rashidi2011activity, rashidi2010multi,wang2018activity}  \newline 
    \textbf{Instance Weighing:} \newline Web-knowledge based Activity Similarity \cite{zheng2009cross, hu2011cross}, Similarity-based \cite{guo2023transferred, hu2011cross} and Boosting \cite{inoue2016supervised} \newline 
    \textbf{Source Domain Selection:} \newline Sensor similarity \cite{yu2023optimal,sonia2020transfer}, $\mathcal{H}$ distance \cite{rashidi2011domain}, Word2Vec \cite{niu2022source} \newline 
    \textbf{Synthetic Data Generation:} \newline User-survey \& Simulation \cite{ali2023user, ali2019improving}, GANs \cite{soleimani2021cross} \newline  
    &
    \textbf{Instance Mapping \& Label Propagation:} \newline Clustering \cite{plugnlearn_rokni_2016,collaborative_kong_2020,recurrenthar_kumar_2022,evaluating_wijekoon_2020,personalized_bettini_2021}, Graph matching \cite{autonomous_rokni_2018,alinia2020actilabel,autonomous_rokni_2018,synchronous_rokni_2017}, Optimal transport \cite{crossdomain_lu_2021}, QR factorization \cite{opportunistic_banos_2021}  \newline 
    \textbf{Instance Weighing:} \newline
    Bagging \& Boosting \cite{online_sztyler_2017,framework_wen_2017,enabling_lane_2011}, Similarity measures \cite{towards_li_2022,feature_chen_2021,importanceweighted_hachiya_2012}, Similarity networks \cite{enabling_lane_2011}, Kernel mean matching \cite{transfer_li_2022} and Nearest neighbor weighing \cite{unsupervised_barbosa_2018}, Misclassification based \cite{swladapt_hu_2023,endtoend_ye_2021,swladapt_hu_2023}
    \newline 
    \textbf{Source Domain Selection:} \newline
    Distance measurements \cite{deep_wang_2018,crossposition_chen_2019,dmstl_zhu_2023,xhar_zhou_2020}, Gating function \cite{sharenlearn_rokni_2019,transfer_shen_2023}
    \newline 
    \textbf{Synthetic Data Generation:} \newline 
    Using skeletal \& MoCap data \cite{novel_fan_2022,learning_xia_2021}, Using videos \cite{kwon2020imutube}, Using text \cite{leng2023generating}, GANs \cite{systematic_chang_2020,generic_soleimani_2022}, Data augmentation \cite{systematic_chang_2020,semanticdiscriminative_lu_2022}
    \newline  
    \\
    
    Feature Space Transfer & 
    \textbf{Conventional:} \newline Sensor profiling \cite{guo2023transferred, sanabria2020unsupervised,chiang2012knowledge,niu2022source,yu2023optimal,van2008recognizing,cook2010learning}, Semantic/Meta features \cite{ye2018slearn,wang2018activity,feuz2015transfer,ye2020xlearn,ye2014usmart,van2010transferring,samarah2018transferring}, Sparse coding \cite{dridi2022transfer}, Random forest \cite{sukhija2019supervised}, Genetic Algorithm \cite{heterogeneous_feuz_2014}, Procrustes analysis \cite{feuz2017collegial} Ontological methods \cite{chen2013ontology, hiremath2022bootstrapping, ye2014usmart}, Similarity functions \cite{chen2017activity,sukhija2018label,guo2023transferred,hu2011transfer,samarah2018transferring} \newline
    \textbf{Neural Network Based:} \newline 
    \textbf{(a) Architectures:} \newline Word2Vec \cite{yu2023fine,niu2020multi,niu2022source}, Neural word embedding \cite{azkune2020cross} \newline
    \textbf{(b) Loss Functions:}
    \newline
    \textbf{(c) Training Strategies:} \newline
    
    &
    \textbf{Conventional:} \newline
    PCA \cite{achieving_lago_2020}, LDA \cite{framework_wen_2017}, Cluster-and-match \cite{personalization_fallahzadeh_2017}, Feature agglomeration \cite{additional_lago_2021}, TCA \cite{unsupervised_barbosa_2018}, SA \cite{unsupervised_barbosa_2018}, LLE \cite{personalized_saeedi_2018}, CORAL \cite{transfer_li_2022}, Sparse coding \cite{bhattacharya2014using}, Matrix factorization \cite{instilling_wei_2016}, Dependency graph mapping \cite{modelagnostic_alinia_2023}, Context models \cite{adaptive_wen_2016}, Semantic space representation \cite{nuactiv_cheng_2013,towards_cheng_2013}
    \newline
    \textbf{Neural Network Based:} \newline
    \textbf{(a) Architectures:} 
    \newline
    \textbf{(b) Loss Functions:} \newline Adversarial domain discrimination/confusion loss \cite{chakma2021activity,systematic_chang_2020,dmstl_zhu_2023}, Contrastive loss \cite{contrasgan_sanabria_2021}, MMD loss \cite{crossposition_chen_2019,deep_wang_2018,dmstl_zhu_2023,feature_chen_2021,multisource_dong_2023}, Triplet loss \cite{domain_alajaji_2023,endtoend_ye_2021}, $L_{2,1}$ Matrix norm loss \cite{feature_chen_2021}, KL divergence loss \cite{scaling_khan_2018}, and Large margin loss \cite{semanticdiscriminative_lu_2022}, JSD Loss \cite{augtoact_faridee_2019}, elf-distillation loss \cite{tasked_suh_2022} \newline
    \textbf{(c) Training Strategies:} \newline
    \\
    
    Parameter Transfer &
    \textbf{Conventional:} \newline Active learning \cite{lu2013hybrid}, Co-training \cite{feuz2017collegial}, Ensemble Learning \cite{ye2020xlearn,rashidi2011activity, rashidi2010multi,heterogeneous_feuz_2014}, HMM \cite{van2010transferring} \newline
    \textbf{Neural Network Based:} \newline \textbf{(a) Architectures:} \newline LSTM \cite{yu2023fine}, AE \cite{rahman2022enabling}, Graph AEs \cite{medrano2019enabling}, Bi-GAN \cite{sanabria2021unsupervised}, DBN \cite{ramamurthy2021star} \newline
    \textbf{(b) Loss Functions:} \newline
    \textbf{(c) Training Strategies:} \newline 
    Adversarial learning \cite{sanabria2021unsupervised}
    \newline
    & 
    \textbf{Conventional:} \newline self-training \cite{crossdomain_thukral_2023}, Co-training \cite{improving_longstaff_2010}, Active learning \cite{improving_longstaff_2010,nuactiv_cheng_2013,online_sztyler_2017,personalizing_akbari_2020}, Knowledge distillation \cite{collaborative_kong_2020}, Ensemble learning \cite{autonomous_rokni_2016, sharenlearn_rokni_2019,crossposition_chen_2019,stratified_wang_2018}, ELMs \cite{crossmobile_zhao_2010} and Decision trees \cite{crosspeople_zhao_2011}, Probabilistic Bayes-optimal classifiers \cite{active_diethe_2016}  \newline
    \textbf{Neural Network Based:} \newline
    \textbf{(a) Architectures:} \newline AE \cite{crossdomain_khan_2022,untran_khan_2018},
    CNN \cite{cnn_chikhaoui_2018,domain_alajaji_2023,human_varshney_2021,scaling_khan_2018,sensehar_jeyakumar_2019,transnet_rokni_2020}, Bayesian CNN \cite{personalizing_akbari_2020}, GRU \cite{deeptranshar_kumar_2023,recurrenthar_kumar_2022}, LSTM \cite{digging_hoelzemann_2020}, CNN-LSTM \cite{deep_wang_2018}, CNN-GRU \cite{deeptranshhar_kumar_2022}, VAE \cite{transferring_akbari_2019,novel_fan_2022}, Bi-GAN \cite{contrasgan_sanabria_2021}, GNN \cite{deep_yan_2022}, and transformers \cite{dmstl_zhu_2023,domain_alajaji_2023,strangan_faridee_2021}
    \newline
    \textbf{(b) Loss Functions:} \newline
    \textbf{(c) Training Strategies:} \newline Adversarial Learning \cite{contrasgan_sanabria_2021,chakma2021activity,systematic_chang_2020}, Self-supervised learning \cite{multitask_saeed_2019, haresamudram2022assessing, haresamudram2023investigating}, Disentangled Learning \cite{learning_xia_2021}, Model agnostic meta-learning \cite{adapting_gong_2022, metasense_gong_2019}, Soft parameter sharing \cite{domain_hussein_2022,metier_chen_2020}, Domain adaptive batch normalization \cite{incremental_mazankiewicz_2020}, Joint training \cite{metier_chen_2020}, Leave-one-dataset-out \cite{combining_presotto_2023}, Using pre-trained models of other modalities \cite{vision_radu_2019,transfer_pavliuk_2023}, Cascaded learning \cite{transfer_du_2019} \newline
    \\
\bottomrule
\end{tabular}
\end{adjustbox}
\end{table}

\subsubsection{More exploration of transfer learning methods for HAR in smart home settings:}
As seen from Table \ref{tab:solution_space}, the application domain of smart homes lacks both the numbers as well as the variety in the transfer learning methods proposed. 
Specifically, we find $61$ works attempting transfer learning in smart homes, whereas, the same for wearable HAR is $143$. 
In terms of the solution space, we note that a direct comparison is not possible since smart homes and wearable HAR domains present unique challenges. 
For example, methods like synthetic data generation using data augmentations are not straightforward in smart homes as the feature spaces across source and target settings are drastically different. 
However, methods based on neural networks for feature space and parameter transfer, that focus on using different neural network architectures, loss functions, and training strategies have been significantly underexplored (see Table \ref{tab:solution_space}).
Exploration of these methods is possible, especially with the use of already-proposed techniques like Word2Vec and neural word embeddings, which are capable of projecting multiple kinds of feature spaces onto a common space.  

\subsubsection{Learning from the state-of-the-art from the Wearable HAR domain:}
As discussed in Section \ref{subsec:review_sota}, challenges like feature space heterogeneity, personalization, multi-source training, etc. are common to both the application domains of smart homes and wearables.
Thus, methods that have been tested to perform best to address challenges in the wearable domain can also be used for performing transfer learning in smart home settings.
Such methods include the use of neural networks based on AE, CNN, GRU/LSTSM, and attention modules, loss functions based on MMD and domain discrimination loss, and training strategies like model-agnostic meta learning, leave-one-dataset-out, soft parameter sharing, self-supervised learning, and disentangled learning. 
We propose building transfer learning frameworks based on these methods as a potential research direction.

\subsubsection{Designing methods for deployable systems:}
The challenges of data security, privacy, and dynamic environments are inevitably present in real-life scenarios, yet, they are rarely addressed in the literature. 
For example, we only find 4 works that perform federated/distributed learning in the wearable HAR literature. 
Also, most of the works in these domains rely on conventional approaches that do not rely on neural networks (See works from Table \ref{tab:smart_home_datasets_challenges}), while most of the state-of-the-art methods heavily focus on them. 
Another challenge in these domains is the lack of consistent evaluation frameworks.
While addressing these challenges, the conditions necessary to simulate the real-world setups are simulated in different ways, using publicly available datasets. 
This calls for designing a standardized evaluation framework to test and compare approaches that address these challenges, which we propose as a future work direction. 

\subsubsection{Development of new transfer learning frameworks:}
Based on Table \ref{tab:wearable_sota}, we see a trend in the recent transfer learning works, where new training strategies are designed. 
While neural network architectures and loss functions are essential components, it is the advances in newer training strategies like self-supervised learning, disentangled learning, and cross-modality transfer that result in achieving transfer learning systems that can be applied to diverse settings and provide performance improvements.  

Under the self-supervised learning paradigm, the design of new pretext tasks suitable for multi-modal, heterogeneous, and sequential data is a well-sought research direction. 
In addition, different contrastive learning based pretext tasks have been explored to \textit{direct} the representation space such that similar data instances are projected close to each other. 
While such different pretext tasks are \textit{designed} to learn different \textit{kinds} of representations, learning what kinds of representations are crucial under transfer for a specific problem, followed by constructing the appropriate pretext task accordingly is also a promising direction for future works.

Achieving disentanglement in the representation space is aimed at learning features that correspond to semantically independent aspects of the data. 
For example, the user-specific independent factors controlling the sensor data generated by a specific individual could be their age, height, and gender.
Disentangling the representation space based on these factors could potentially result in a system that can be adapted for a new user that differs in a subset of these factors.
By performing \textit{instrumentation} on this disentangled representation space, changing one of these factors for this new user would not affect the data generated from the other common factors.
We propose future research in this domain along the lines of investigating such independent factors and building frameworks to disentangle them. 

Cross-modality transfer is yet another exciting and promising field, where the aim is to leverage larger data amounts and existing pre-training models present in other modalities like vision and natural language.
Under this paradigm, recent works have mainly focused on directions like synthetic sensor data generation \cite{kwon2020imutube, leng2023generating} and performing inference on sensor data using pre-trained models \cite{transfer_pavliuk_2023}.
Collecting large-scale sensor datasets that cover a large number of heterogeneities as mentioned in Section \ref{sec:smart_home} and Section \ref{sec:wearables} is non-pragmatic. 
Therefore, generating such sensor data virtually is an exciting avenue to address the data scarcity problem. 
In addition to this, another breakthrough has been achieved \cite{girdhar2023imagebind}, where a common representation space is learned for multiple modalities. 
Learning such a representation space is especially useful for tackling the problem of diverse feature spaces encountered in typical HAR settings. 
We propose that the HAR community can benefit by investing efforts in these research directions.

%% file: conclusion.tex
The biggest challenge faced by the HAR community is the labeled data scarcity that covers a range of target settings. 
Variation in these settings in terms of sensor modalities, positions, subject characteristics, label spaces, etc., induces a domain shift that is detrimental to the adaptive performance of HAR models trained in a particular setting. 
The solution to this problem lies in transfer learning, where the goal is to reuse the knowledge gained in one/more source domain(s) under the presence of domain shift. 
In this survey, we analyzed $205$  papers that describe transfer learning in the HAR settings. 
Specifically, we focused on the application domains of smart homes and wearable HAR and selected prominent works that represent a wide range of transfer learning methodologies. 

Our survey highlights that addressing feature space heterogeneity, personalization, and task differences simultaneously is extremely rare in the transfer learning literature. 
These challenges are present in many real-life source-target settings, which must be addressed by the upcoming works in this domain. 
Our analysis also indicates that a significantly lesser amount of exploration has been done for performing transfer learning in smart home settings compared to the wearable HAR domain.
In addition to these gaps in the literature, we also point out the promises offered by some of the recent trends like self-supervised learning, disentangled learning, and cross-modality transfer, which we believe are crucial for taking the next steps toward solving the problem of HAR. 
We hope that this survey stands as a reference point for a researcher to: (\textit{i}) get accustomed to the transfer learning methods used in the field of HAR and (\textit{ii}) understand the relevant challenges and possible ways of addressing them.

%% file: appendix.tex
\subsection{Survey Methodology}

\begin{figure}[]
  \centering
  \includegraphics[width=\linewidth]{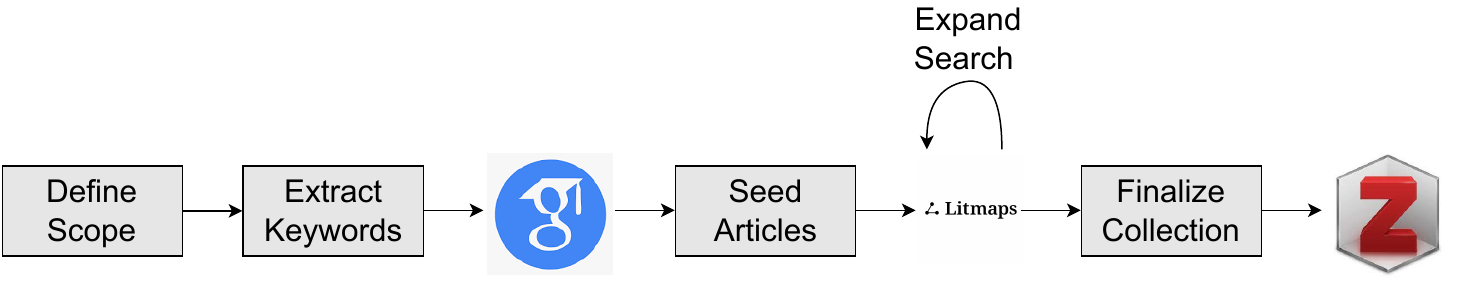}
  \caption{Illustration of the survey methodology: To conduct a comprehensive survey, we make use of literature review tools like Litmaps and Zotero.
  Litmaps offers functionalities like article seeding, discovery, and visualizations that are particularly helpful for covering all the relevant works in the selected domain, whereas, Zotero offers storage and categorization capacities. }
  \label{fig:survey_method}
\end{figure}

Here, we briefly describe the methodology employed to perform our literature survey. 
We first defined the scope of our survey, which is presented in section \ref{sec:introduction}. 
Then, we extracted the keywords that: (\textit{i}) are likely to be present in our target articles and (\textit{ii}) represent the contribution of our intended collection. 
Our keywords include "\textit{activity recognition}", "\textit{action recognition}", "\textit{HAR}", "\textit{transfer learning}","\textit{domain adaptation}", "\textit{cross-domain}", "\textit{cross-modality}", "\textit{cross-subject}", "\textit{cross-position}", "\textit{domain invariant}", "\textit{new environment}", etc. 
Then, we considered a meaningful combination of these keywords and used them to search for relevant articles on Google Scholar. 
The intention behind this step was to collect only a sample of prominent contributions like the survey and well-cited/recognized papers in the field. 
After collecting a set of $\sim15-20$ such articles, we imported it on a research discovery tool called Litmaps \cite{litmaps} and expanded the search using its Discover functionality.   
The Discover tool uses factors like common references, citations, and keyword similarity to search for articles that are similar to the imported seed collection. 
Thus, the Discover tool acts as a primary level of screening to recommend articles that are related to our seed collection. 
Then, we manually checked the recommended articles to make sure they lie inside our defined scope, which we consider as the secondary level of screening. 
Next, we added the newly recommended articles that passed through our manual check into our global collection and expanded the search by iteratively performing the said procedure. 
We performed such iterations until Litmaps started providing incorrect recommendations, which we considered as an indicator of an exhaustive search. 
Finally, we went through the resultant collection one more time to verify that the articles present were within the bounds of our scope. 

\subsection{Background on Transfer Learning Approaches}
Instance transfer refers to a methodology that reuses the data instances from the source domain $X^s$ in solving the target domain task $T^t$. 
The traditional interpretation of this approach is to re-weight the data points based on a similarity/importance metric and transfer them in the learning process of the target domain. 
However, weighting of data points is not the only way in which $X^s$ can be reused in performing $T^t$.
The implicit use of $X^s$ towards improving $T^t$ can be witnessed in approaches like source domain selection, label space transfer, and synthetic data generation. 
In source domain selection, importance scores are assigned to the entire domains rather than individual instances, in a multi-source setting.
This allows one to select a more compatible set of source domains for a given target domain.
Label space transfer takes the approach of assigning $X^s$ a label from the set $Y^t$ (or vice versa) by learning a mapping between $Y^s$ and $Y^t$.  
This ensures that source instances can also be used in target domain training in a supervised manner, owing to the newly assigned target domain labels.
In synthetic data generation, the idea is to leverage data instances from source domain(s) and their properties for generating artificial data instances for the target domain. 

Feature transfer is a type of transfer learning method, where both the source and the target data instances are mapped onto a common feature space.
The intuition behind it is to enable the source classifier to perform inference on the target domain after the data instances are mapped accordingly. 
The common feature space can simply be the target space, in which case the source data can be mapped to the target feature space (or vice versa) and a common model can be learned. 
Another way is to learn an intermediate feature space, different from both source and target domains, which can be \textit{designed} in a way that boosts the transfer learning performance. 
In the literature, two themes of such feature transfer approaches are found: (\textit{i}) conventional algorithmic approaches and (\textit{ii}) deep learning based approaches. 
Conventional approaches learn the required mapping function by optimizing various statistical measures associated with the distribution of data instances in a closed form solution.
On the other hand, deep learning based approaches model the required mapping function using neural networks, which are trained to optimize distribution specific characteristics of the mapped data instances. 

The basis of parameter transfer lies in the assumption that a model trained on $T^s$ is capable of learning both setting-specific and setting-invariant information. 
Thus, the goal here is to transfer the parameters learned in the source setting, which are mostly setting-invariant, so that they can assist the $T^t$. 
While the traditional approaches to performing parameter transfer include methods like self-training, co-training, active learning, pseudo labeling, and ensemble learning, the recent contributions performing parameter transfer leverage neural networks, where it is easier to learn and transfer only specific parameters across domains. 
This can be done by customizing the neural network training in terms of architecture, training strategies, and setups.

While many prominent works \cite{pan2009survey, weiss2016survey} classify transfer learning into instance, feature, parameter, and relational-knowledge transfer, the latter is applicable only for relational-domains, which is rare in sensor-based HAR datasets. 
Rather, literature consists of works where \textit{insights} from domains are transferred across to the target domain. 
These insights consist of domain-specific metadata, heuristics, and context information, which often help in designing the frameworks to perform the target domain task.
Thus, we classify such approaches under knowledge-base transfer, which is more relevant in case of transfer learning in sensor-based HAR.  
Table \ref{tab:TL_approaches} summarizes this classification system from the solution perspective and highlights important themes, which are discussed above. 

\begin{figure}[]
  \centering
  \includegraphics[width=0.72\linewidth]{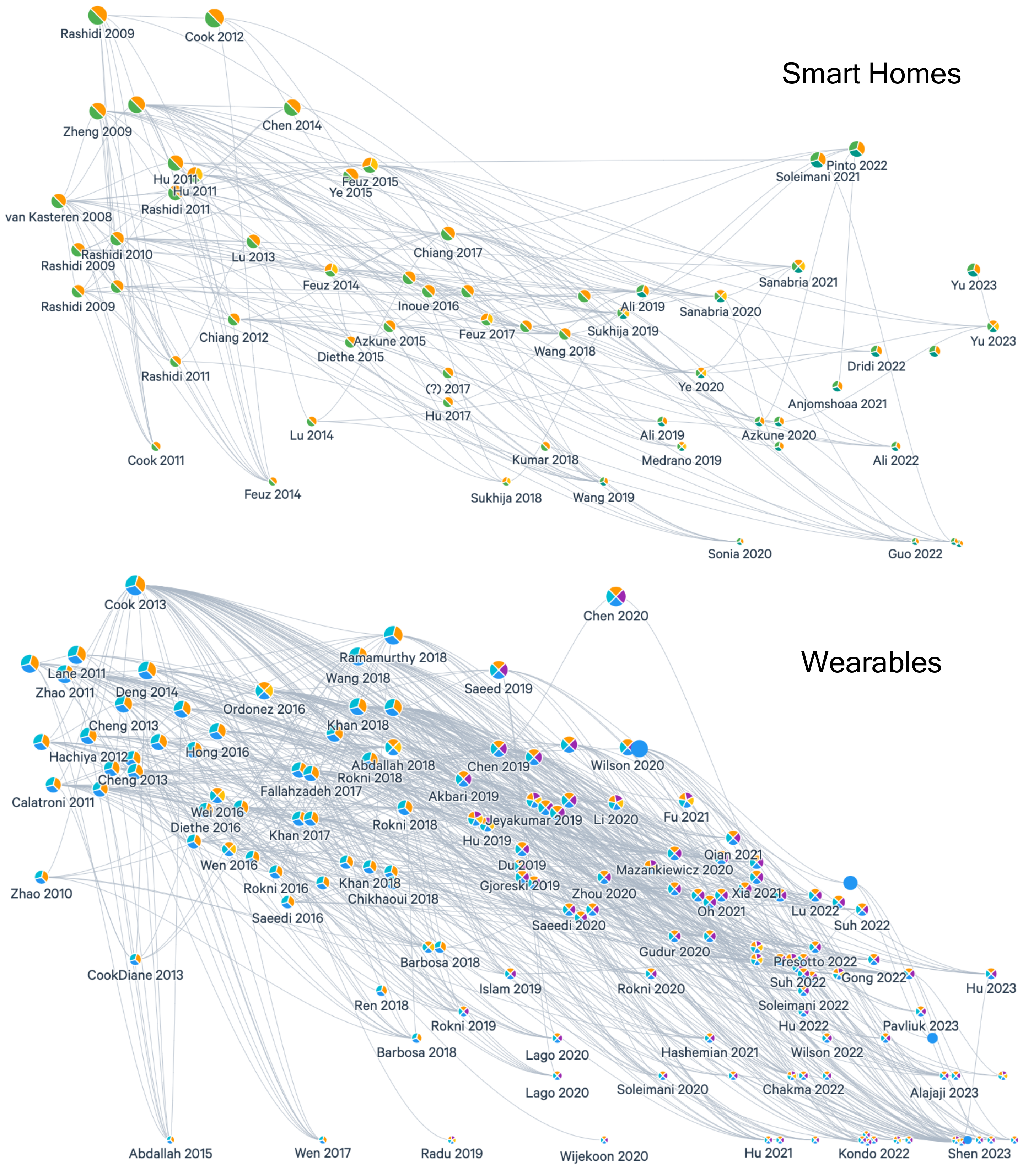}
  \caption{Literature Map of transfer learning works performing HAR in the application domain of smart homes (top) and wearables (bottom): 
  In this map, the X-axis denotes the year of publication, whereas, the Y-axis denotes the number of citations. 
  This map is helpful for: (\textit{i}) verifying if the literature review is comprehensive, (\textit{ii}) filtering out the recent as well as prominent (well-cited) works, and (\textit{iii}) analyzing the trends in recent past.
  These maps are generated using Litmap.}
  \label{fig:litmap}
\end{figure}

\subsection{Litmap Visualizations (Figure \ref{fig:litmap})}
To conduct our survey, we use an online tool called Litmaps \cite{litmaps}. 
It provides a visualization functionality, where one can examine the spread of the manuscripts present in the survey in terms of their publication year and number of citations. 
These visualizations can also be used for: (\textit{i}) verifying if the literature review is comprehensive, (\textit{ii}) filtering out the recent as well as prominent (well-cited) works, and (\textit{iii}) analyzing the trends in recent past.
The survey collection is more likely to be comprehensive when the constituent papers in it are well-connected in the map visualization.
Thus, for the papers that are sparsely connected in the map, we manually check for related articles by going through their citations on Google Scholar. 
For filtering the recent as well as relevant works, we check the papers that lie on the top (more cited) and right (more recent) directions of the map.    

